\documentclass[runningheads]{llncs}

\usepackage{eccv}

\usepackage{eccvabbrv}

\usepackage{graphicx}
\usepackage{booktabs}

\usepackage[accsupp]{axessibility}  

\usepackage{multirow}
\usepackage{algorithm}
\usepackage{algpseudocode}
\usepackage{colortbl}
\usepackage{xcolor}
\usepackage{amsmath}
\usepackage{pifont} 
\usepackage{calrsfs} 
\usepackage{marvosym}
\usepackage{subcaption}
\usepackage[export]{adjustbox}
\DeclareMathAlphabet{\pazocal}{OMS}{zplm}{m}{n}
\let\originalllncsparagraph\paragraph
\let\originalllncssubparagraph\subparagraph

\let\paragraph\undefined
\let\subparagraph\undefined

\usepackage{titlesec}
\titlespacing*{\section}{0pt}{0.4ex}{0.1ex}
\titlespacing*{\subsection}{0pt}{0.3ex}{0.1ex}
\titlespacing*{\subsubsection}{0pt}{0.2ex}{0.1ex}

\let\paragraph\originalllncsparagraph
\let\subparagraph\originalllncssubparagraph

\usepackage{hyperref}

\usepackage{orcidlink}

\begin{document}

\title{Syn2RealTrack: Bridging the Gap Between Synthetic and Real-World Datasets for Online Multi-View Multi-Target Tracking}

\titlerunning{Syn2RealTrack: Online MTMC Tracking}

\author{Duong Nguyen-Ngoc Tran$^*$\orcidlink{0000-0001-7537-6377} \and
  Ngoc Doan-Minh Huynh$^*$\orcidlink{0000-0003-0662-4036} \and
  Cu Quoc Le$^*$\orcidlink{0009-0005-9617-1983} \and
  Khang Nguyen Hoang$^*$\orcidlink{0000-0003-3616-0367} \and
  Long Hoang Pham\orcidlink{0000-0002-3240-657X} \and
  Huy-Hung Nguyen\orcidlink{0000-0001-5394-9381} \and \\
  Quoc Pham-Nam Ho\orcidlink{0009-0006-7256-4798} \and Trinh Le Ba Khanh\orcidlink{0000-0003-1688-9003} \and
  Chi Dai Tran\orcidlink{0000-0001-6345-8658} \and 
  Duong Khac Vu\orcidlink{0009-0009-4715-0704} \and 
  Son Hong Phan\orcidlink{0009-0000-0572-6685} \and 
  Hyung-Min Jeon \and
  Jae Wook Jeon$^\vartriangle$\orcidlink{0000-0003-0037-112X}}

\authorrunning{D. N.-N. Tran et al.}

\institute{Department of Electrical and Computer Engineering, \\
  Sungkyunkwan University, Suwon, South Korea \\
\email{\{duongtran, ngochdm, lequoccu2003, khangnguyen, jwjeon\}@skku.edu}}

\maketitle

\def\thefootnote{*}
\footnotetext{These authors contributed equally to this work.}
\def\thefootnote{$^\vartriangle$}
\footnotetext{Corresponding author.}
\def\thefootnote{\arabic{footnote}}

\begin{abstract}
  Multi-camera 3D perception systems for warehouse scenes are trained largely on synthetic data and evaluated on physically captured environments. The resulting synthetic-to-real gap, which corrupts ground-plane localization and cross-camera identity association, is usually treated as one deficiency for a single domain-adaptation module to absorb; we argue instead that it enters the pipeline at three separable points: the camera calibration, the object shape prior, and the assumption that the object census is known, each admitting a different local remedy. Our online pipeline, Syn2RealTrack, follows this decomposition: lens distortion is recovered from images alone under a calibration that provides none, detections are fused across views by a visibility-weighted part-based descriptor that abstains on occluded parts rather than guessing, person height is measured in closed form from calibration instead of copied from a synthetic prior, and a closed-world cardinality prior is paired with a causal filter that removes the phantom boxes the prior manufactures. The system therefore adapts by reallocating trust between geometry and appearance without retraining a feature extractor. On the AI City Challenge 2026 Track~1 evaluation server it reaches a 3D Higher Order Tracking Accuracy (HOTA) of $52.0118\%$. The code will be released at \url{https://github.com/SKKUAutoLab/aic26_mc3dp}.
\end{abstract}

\section{Introduction}
\label{section:introduction}

\begin{figure}[t]
	\centering
	\includegraphics[width=0.98\linewidth,valign=c]{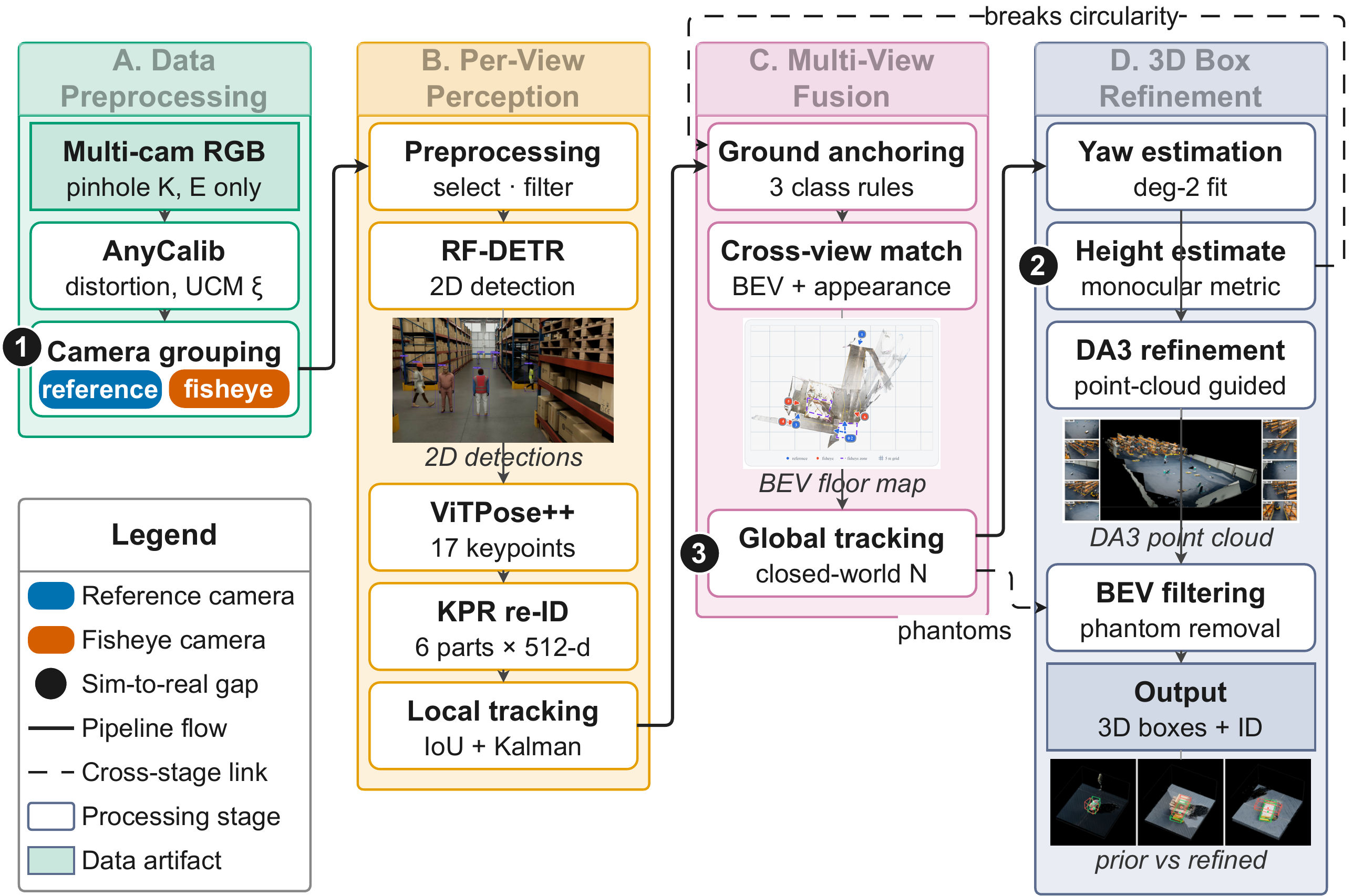}
	\caption[Overview of the proposed framework.]{Overview of the proposed framework, read left to right. \textbf{(A)~Data preprocessing.} AnyCalib~\cite{tirado-garin_anycalib_2025} estimates per-camera distortion from $31$ sampled frames, and the Unified Camera Model (UCM) parameter $\xi$ separates reference ($\xi \le 0.3$, blue) and fisheye-candidate (red) views. \textbf{(B)~Per-view perception.} RF-DETR~\cite{robinson2026rfdetr}, ViTPose++~\cite{xu_vitpose_2024}, and Keypoint Promptable Re-Identification (KPR)~\cite{DBLP:conf/eccv/SomersAV24} provide detections, keypoints, and visibility-aware part descriptors, which an appearance--IoU Kalman tracker links into local identities. \textbf{(C)~Multi-view fusion.} Class-specific ground anchors and visibility-weighted part similarity merge observations within per-class bird's-eye-view (BEV) gates; local-identity carry-forward and gated Hungarian assignment then form global tracks under cap $N_k$. \textbf{(D)~3D box construction and refinement.} Trajectory-derived yaw, monocular person height, footprint and yaw refinement guided by Depth Anything~3 (DA3)~\cite{lin2026depth}, and BEV visibility filtering produce the final boxes. Numbered badges identify the three synthetic-to-real gaps (calibration, shape priors, and closed-world cardinality), and dashed links show cross-stage dependencies between measured height, ground anchoring, and BEV filtering.}
	\label{fig:main_framework}
	\vspace{-0.8cm}
\end{figure}

Warehouse automation depends on knowing where every person and every piece of equipment is, in metric floor coordinates, at every instant. A single camera cannot provide this information: it recovers a bearing but not a range, and, among tall racks, a target may disappear behind an obstruction within a few frames. Fixed multi-camera rigs with overlapping fields of view are the standard remedy, and the task they define is multi-target multi-camera (MTMC) 3D perception: recovering each object's world-frame position, footprint, and orientation, together with an identity that persists through occlusions and handovers between cameras~\cite{tang_9th_2025}. The obstacle is supervision. Dense 3D boxes and cross-camera identities are expensive to annotate at scale in a working facility~\cite{woo_mtmmc_2024}, so the practical solution has been to train in simulation, where exact ground truth is freely available. What simulation does not capture is the collection of small physical irregularities that accumulate in a real installation: a lens that does not conform to the calibration on file, an object whose true dimensions differ from those in the catalogue, and a floor whose occupancy has not been counted.
\begin{figure}[t]
    \centering
    \begin{minipage}{0.98\linewidth}
        \centering
        \includegraphics[width=\linewidth,valign=c]{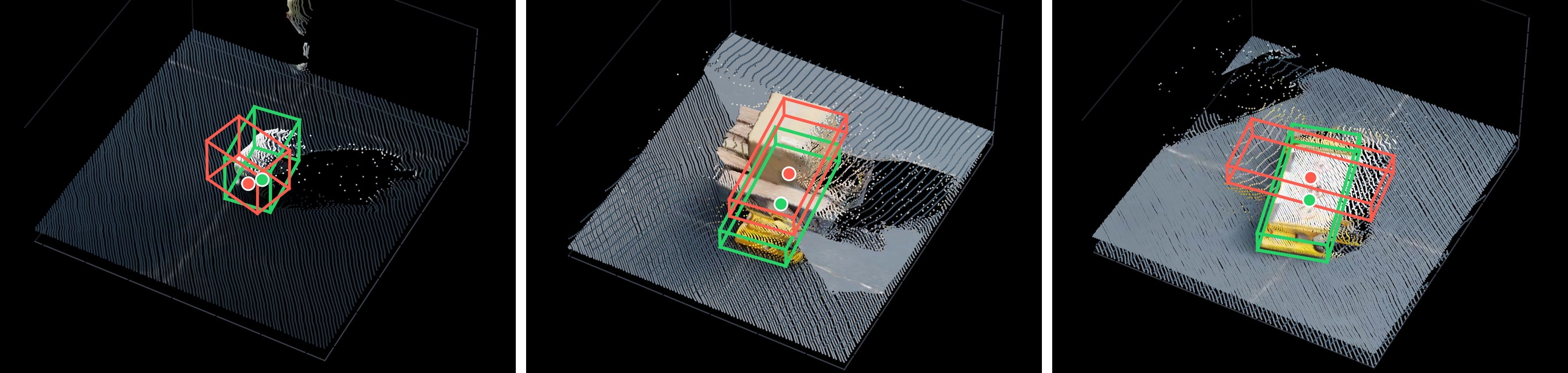}\\
        \scriptsize (a) Synthetic scene
    \end{minipage}
    \begin{minipage}{0.98\linewidth}
        \centering
        \includegraphics[width=\linewidth,valign=c]{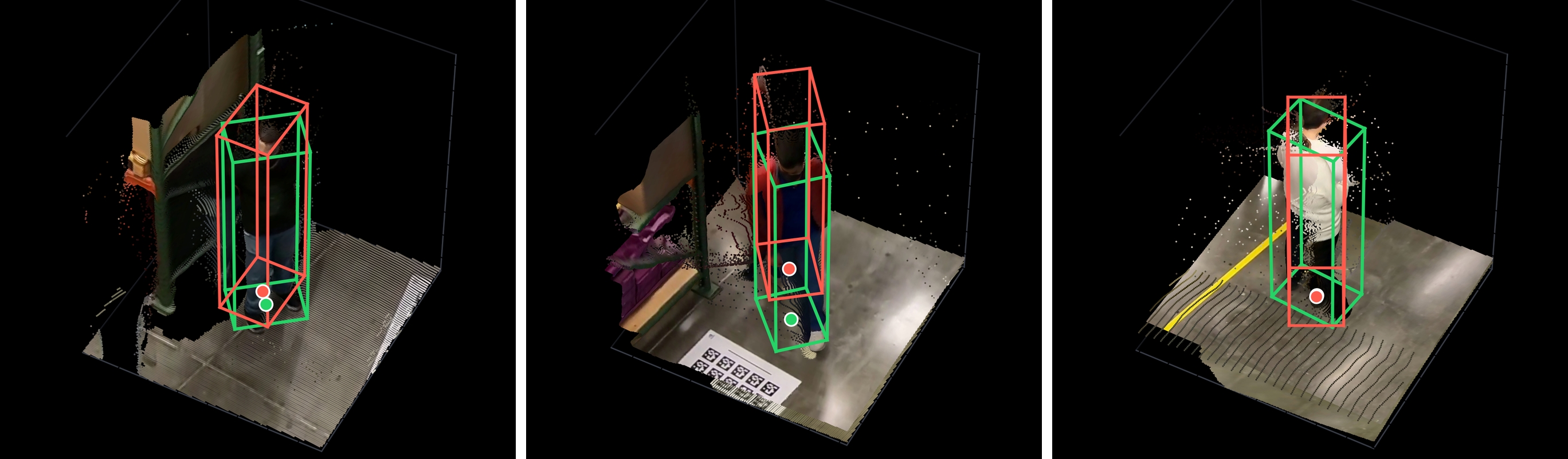}\\
        \scriptsize (b) Real-world scene
    \end{minipage}
	\vspace{-0.4cm}
    \caption{Representative point-cloud-guided 3D box refinement examples on (a) a synthetic and (b) a real-world scene of the AI City Challenge 2026 Track~1 dataset~\cite{Tang26AICity26}. Red boxes denote the original multi-view tracking prior, green boxes denote the refined 3D boxes, and the colored points denote the local cloud reconstructed by Depth Anything~3 (DA3)~\cite{lin2026depth}.}
    \label{fig:point_cloud_guided_refinement_examples}
	\vspace{-0.6cm}
\end{figure}
We present Syn2RealTrack, our AI City Challenge 2026 Track~1 system: an online multi-camera 3D perception pipeline trained mainly on synthetic data and evaluated partly on real scenes~\cite{Tang26AICity26}; \cref{fig:point_cloud_guided_refinement_examples} illustrates the two domains and our point-cloud-guided box refinement on both. Rather than treating the synthetic-to-real gap as one domain-adaptation problem, we localize it to three interfaces (camera calibration, object-shape priors, and known object counts) and address each where it arises. \cref{fig:main_framework} summarizes the framework; our contributions are as follows.
\begin{itemize}
  \item \textbf{Localizing the synthetic-to-real gap.} With feature extractors trained once and then frozen across domains, we expose calibration, camera overlap, and object counts as a configuration surface that shifts trust from simulation-specific geometry to appearance when these cues are unreliable in real scenes (\cref{tab:sim2real_surface}).
  \item \textbf{Distortion-aware camera grouping.} As the calibration omits lens distortion, we estimate it with AnyCalib~\cite{tirado-garin_anycalib_2025}, split reference and fisheye views by the Unified Camera Model parameter $\xi$, undistort footpoints while keeping the provided geometry, and mask fisheye-only zones to avoid invalid matches.
  \item \textbf{Cross-view fusion that can abstain.} We ground detections, fuse views by visibility-weighted part similarity, and propagate single-camera identities before geometric assignment to reject inconsistent world placements; this fusion adds $0.06$ HOTA on top of single-view association.
  \item \textbf{Measured rather than synthetic geometry.} For people, a closed-form estimate from calibration and ankle ground points replaces the synthetic height prior without depth or extra supervision; an RGB-only Depth Anything~3 (DA3) point cloud~\cite{lin2026depth} then refines footprint and yaw while preserving identity and class-specific size.
  \item \textbf{A contained closed-world prior.} We apply exact per-class counts only in suitable closed-world scenes and suppress unsupported boxes with a causal filter requiring visible-ankle confirmation after reprojection into covering cameras; this filter yields a detection-driven gain at negligible association cost.
\end{itemize}

\section{Related Work}
\label{section:related_works}

\noindent\textbf{Synthetic-to-Real Generalization and Calibration.} Multi-camera domain shift is both photometric and geometric. Prior work improves transfer through view-consistent augmentation, depth- and pose-robust BEV features, or perspective-invariant rendering~\cite{engilberge_augmentation_2023,wang_dg_bev_2023,lu_perspective_rendering_2025,huynh_tsbow_2026}; calibration methods instead recover camera geometry from person correspondences, learned perspective cues, or predicted pixel rays~\cite{xu_widebaseline_calibration_2021,veicht2024geocalib,tirado-garin_anycalib_2025,tran_region-and-trajectory_2021}. Our pipeline keeps its feature extractors and the provided intrinsics and extrinsics fixed; it adapts only at explicit downstream interfaces and estimates each camera's omitted lens distortion before ground-plane projection.

\noindent\textbf{Geometry, Visibility, and Multi-View Association.} Prior work localizes people and recovers metric shape through calibrated pose, projective metrology, RGB geometry, or point-supported box refinement~\cite{lima_generalizable_2021,criminisi_single_2000,wang_vggt_2025,ma_dchm_2025,lin2026depth,shi_pointrcnn_2019}. Association methods use visible-part features~\cite{somers_bpbreid_2023,DBLP:conf/eccv/SomersAV24}, probabilistic occupancy, target-count, and camera models~\cite{vo_cphd_2007,ong_bayesian_2022,pham_data_2025}, or learned BEV and trajectory relationships~\cite{teepe_earlybird_2024,wang_mcblt_2025,yamane_mvtrajecter_2025}. Our causal pipeline combines these ideas: class-specific anchors and ankle keypoints recover ground contact and person height; RGB point clouds refine footprint and yaw; shared visible parts, local-identity memory, motion, and a bounded gallery guide sequential association. An external class census conditions assignment when available; otherwise, adaptive tracking and a camera-coverage check suppress unsupported person boxes. The key difference from these works is that our association can abstain: parts occluded in either view are excluded from similarity instead of imputed, and single-camera identities take precedence over geometric matching.

\noindent\textbf{Multi-View Fusion and Tracking.} Multi-view detectors aggregate projected features in BEV using convolution, transformers, or multi-height homographies, while CaMuViD exchanges features directly across uncalibrated views~\cite{hou_mvdet_2020,hou_mvdetr_2021,song_shot_2021,daryani_camuvid_2025,tran_robust_2022}. AI City 2025 systems similarly combine global point-cloud detection, cluster-to-tracklet mapping, or late multi-view tracking and refinement~\cite{tang_9th_2025,lee_pointcloud_reid_2025,tran_depthtrack_2025,phan_vgcrtrack_2025}, but rely on benchmark-provided depth and calibration. The 2026 protocol adds real scenes and withholds depth at inference~\cite{Tang26AICity26}. Our pipeline therefore preserves modular per-view perception and fuses sparse object observations late rather than learning dense BEV aggregation.

\section{Methodology}
\label{section:methodology}

We address online 3D multi-view multi-target tracking in warehouse scenes~\cite{Tang26AICity26,tran_depthtrack_2025}, each observed by $C$ static, calibrated RGB cameras over frames $t \in \{0,\dots,T-1\}$ across seven classes $\mathcal{K} = \{0,\dots,6\}$ (Person, Forklift, NovaCarter, Transporter, FourierGR1T2, AgilityDigit, PalletTruck). Per frame we emit oriented, class-labeled 3D boxes with time- and view-consistent identities. The synthetic-to-real gap surfaces at calibration, the shape prior, and the closed-world assumption.

\noindent\textbf{Notation.}
Camera $c \in \{1,\dots,C\}$ has intrinsics $\mathbf{K}_c \in \mathbb{R}^{3\times3}$ and world-to-camera extrinsics $\mathbf{E}_c = [\mathbf{R}_c \mid \mathbf{t}_c] \in \mathbb{R}^{3\times4}$. An image pixel $\mathbf{u} = (u,v)^{\!\top}$ has homogeneous form $\tilde{\mathbf{u}} = (u,v,1)^{\!\top}$; $\mathbf{x} = (X,Y)^{\!\top}$ is a metric world point on the ground plane $Z=0$. A detection carries a normalized center-size box $\mathbf{b} = (b_x, b_y, b_w, b_h)$, confidence $s \in [0,1]$, class $k \in \mathcal{K}$, $17$ COCO~\cite{lin2014coco} keypoints $\{(\mathbf{u}_j, s_j)\}_{j=0}^{16}$ for humanoid classes, a part-based appearance descriptor $\mathbf{F} \in \mathbb{R}^{6\times512}$ whose rows $\mathbf{f}_p$ are $\ell_2$-normalized per body part, and per-part visibility $\boldsymbol{\nu} \in [0,1]^6$.

\noindent\textbf{Ground-plane projection.}
A ground point has $Z=0$, so the third column of the world-to-image matrix $\mathbf{P}_c = \mathbf{K}_c\mathbf{E}_c \in \mathbb{R}^{3\times4}$ drops out of the projection. For $\bar{\mathbf{P}}_c \in \mathbb{R}^{3\times3}$ formed by columns $1$, $2$, and $4$ of $\mathbf{P}_c$, the ground-plane projection is $\lambda\tilde{\mathbf{u}} = \bar{\mathbf{P}}_c(X,Y,1)^{\!\top}$ at projective depth $\lambda \neq 0$, inverted by the homography $\mathbf{H}_c \triangleq \bar{\mathbf{P}}_c^{-1}$:
\vspace{-0.4cm}
\begin{equation}
  \mathbf{q} \;=\; \mathbf{H}_c\,\tilde{\mathbf{u}},
  \qquad \pi_c(\mathbf{u}) \;=\; \left(\tfrac{q_1}{q_3},\; \tfrac{q_2}{q_3}\right)^{\!\top},
  \label{eq:backproj}
  \vspace{-0.4cm}
\end{equation}
with homogeneous $\mathbf{q} = (q_1,q_2,q_3)^{\!\top}$ and ground position $\pi_c$. We reject back-projections with $|q_3| < \varepsilon_{\text{h}}$, where $\varepsilon_{\text{h}} = 10^{-9}$ (the pixel lies on or beyond the horizon, where its viewing ray never meets the ground), or with $\|\pi_c(\mathbf{u})\|_2 > 10^3\,\text{m}$, which bounds the world extent and discards numerically degenerate rays.

\subsection{Distortion-Aware Camera Grouping}
\label{subsection:distortion_aware_camera_grouping}
\begin{figure}[t]
	\centering
	\includegraphics[width=0.85\textwidth,valign=c]{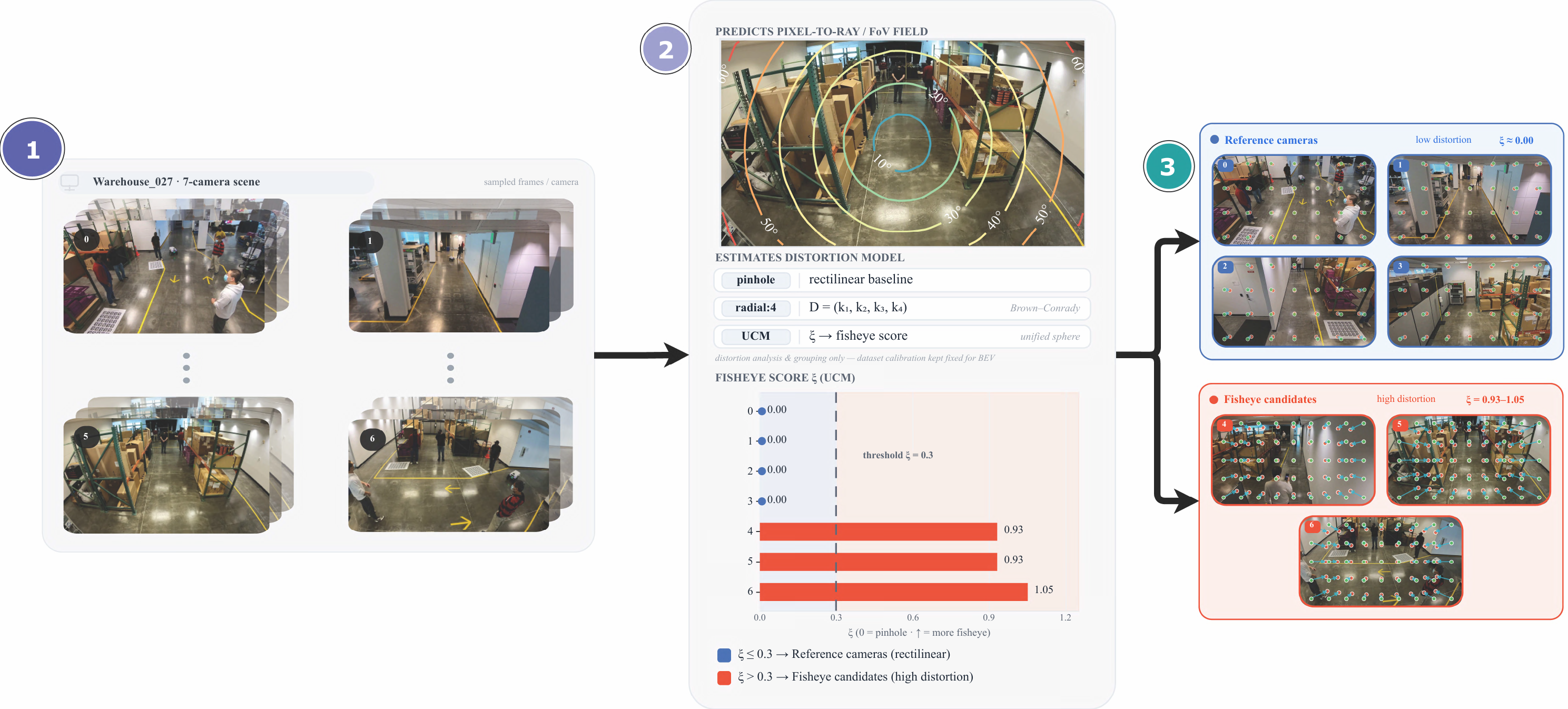}
	\caption{Single-view distortion estimation and camera grouping. (1)~Frames are evenly sampled from every camera of a scene. (2)~AnyCalib~\cite{tirado-garin_anycalib_2025} predicts a dense per-pixel ray and field-of-view map and fits a pinhole baseline, a radial Brown--Conrady model, and the Unified Camera Model (UCM), whose parameter $\xi$ is used as a fisheye score; the estimates are used for distortion analysis and grouping only, and the dataset calibration is kept fixed for bird's-eye-view (BEV) projection. (3)~Cameras with $\xi$ below the threshold become reference views (blue), and the rest become fisheye candidates (red), whose reprojection displacements grow toward the image borders.}
	\label{fig:distortion_grouping}
	\vspace{-0.6cm}
\end{figure}
The dataset calibration~\cite{Tang26AICity26} is pinhole-only: intrinsics, extrinsics, and a ground-plane homography, but no distortion coefficients. With noticeable lens distortion, projected foot points land at the wrong BEV positions, and cross-view association suffers. This step identifies the offending cameras and pulls the projected locations back (\cref{fig:distortion_grouping}). We use AnyCalib~\cite{tirado-garin_anycalib_2025} only for distortion analysis, grouping, and initialization; the dataset pinhole calibration remains the reference geometry for BEV projection.

\noindent\textbf{Single-View Distortion Estimation} (steps 1--2 in~\cref{fig:distortion_grouping}):
We run pretrained AnyCalib on 31 evenly sampled frames per camera and take the per-parameter median. It predicts a dense pixel-to-ray field and fits a pinhole baseline, a radial Brown--Conrady model~\cite{brown_decentering_1966}, and the Unified Camera Model (UCM)~\cite{vernon_unifying_2000,mei_single_2007}; we use the UCM parameter $\xi$ as a fisheye score.

\noindent\textbf{Camera Grouping} (step 3):
Cameras with $\xi \le 0.3$ are reference cameras; the rest are fisheye candidates. This distortion-based split (not a physical lens label) agrees with the estimated radial coefficients and is confirmed by reprojection quiver plots: displacements stay small for reference cameras and grow toward the borders for fisheye candidates.

\subsection{2D Object Detection}
\label{subsection:2d_object_detection}

Because noisy 2D annotations can slow convergence, we classify each category as static, fixed-shape, or dynamic-shape from its per category dimension statistics, providing fallback shapes when matching fails, and remove redundant, tiny, unseen, or invalid boxes. Training scenes are matched to each test scenario, including the otherwise absent `AgilityDigit' and `FourierGR1T2' categories, with real scenes drawn from AI City~\cite{tang_9th_2025} and MTMMC~\cite{woo_mtmmc_2024}.
We train the RF-DETR 2x-large detector~\cite{robinson2026rfdetr} for 200 epochs at $1920$~pixels (batch size 4). Inference uses the same resolution with NMS at IoU~0.5 and a 0.1 confidence threshold. Further details are in the supplementary material (\cref{sec:2D_object_detection_details}).

\subsection{Single-View Re-Identification and Tracking}
\label{subsection:single_view_tracking}

For each detected person, we estimate body keypoints and use them to extract pose-aware descriptors for identity association. Within each camera, motion and these appearance cues then link detections over time into local tracklets, before any cross-view reasoning.

\noindent\textbf{Pose Estimation.} We apply pretrained ViTPose++~\cite{xu_vitpose_2024} to each detection crop. Its transformer encoder and lightweight decoder provide keypoints under changes in pose, scale, and occlusion.

\noindent\textbf{Re-Identification.} Keypoint Promptable Re-Identification (KPR)~\cite{DBLP:conf/eccv/SomersAV24} uses these keypoints (positive only) to split the body into six regions and extract region-level descriptors, reducing background interference and occlusion sensitivity. As identities are scene-specific in AI City 2026 Track~1, we merge and relabel the train/val splits into 155 identities and train KPR for 110 epochs ($\approx$3\,h\,7\,min). These descriptors feed both the single-view tracker below and the multi-view associator of~\cref{subsection:multi_view_tracking}; further details are given in~\cref{sec:re_id_pose_estimation_details}.

\noindent\textbf{Single-View Tracking.} Following tracking-by-detection~\cite{aharon2022bot}, we pair an appearance-augmented IoU tracker with a constant-velocity Kalman filter~\cite{kalman1960new}: box overlap and predicted motion supply the association affinity, and the part-based re-identification features disambiguate crossings and brief occlusions.
Each surviving tracklet gets a local identity $\ell$, and per camera and frame the stage emits instances $(\ell, k, \mathbf{b}, s, \{(\mathbf{u}_j,s_j)\}, \mathbf{F}, \boldsymbol{\nu})$. The global associator of~\cref{subsection:multi_view_tracking} consumes these local identities as a carry-forward cue, since single-camera identity proved more reliable than cross-frame geometric matching alone.

\subsection{Multi-View Tracking}
\label{subsection:multi_view_tracking}

Given the per-camera tracklets, multi-view tracking fuses observations of the same object across cameras into a globally consistent identity, in three steps: each detection is anchored to a ground-plane position by a class-adaptive rule, the resulting observations are merged across views per frame, and the merged world observations are bound to persistent global tracks.

\subsubsection{3.4.1 Class-Adaptive Ground-Contact Anchoring.}

Back-projection through $\mathbf{H}_c$ is exact only for a pixel that genuinely lies on the floor, so the accuracy of the entire bird's-eye-view (BEV) representation reduces to choosing that pixel. A single rule cannot serve all seven classes: a person's feet are visible and semantically well defined, a low, flat robot has no meaningful feet, and a tall vehicle seen from a steep overhead angle has its base occluded by its own body. We therefore select an anchoring strategy per class through a configurable map, with three strategies.

\noindent\textbf{Skeleton.}
For humanoid classes with a reliable pose, the anchor is the midpoint of the two ankle keypoints (COCO indices $15$ and $16$) whenever at least one clears the ankle-confidence threshold $\theta_{\text{ank}}$. When neither does, we extrapolate down the leg's kinematic chain as $\mathbf{u}_{\text{ankle}} = \mathbf{u}_{\text{knee}} + \rho\bigl(\mathbf{u}_{\text{knee}} - \mathbf{u}_{\text{hip}}\bigr)$ with $\rho=1$, where $\mathbf{u}_{\text{hip}}$ and $\mathbf{u}_{\text{knee}}$ are the hip and knee keypoints of one leg and $\rho$ is the shank-to-thigh length ratio, taken as unity for a straight leg; the anchor is the mean over whichever legs admit the construction. This extrapolation fires only when more than $n_{\text{kp}}$ keypoints clear the pose-confidence threshold $\theta_{\text{pose}}$, so that a fragmentary skeleton never fabricates a ground contact.

\noindent\textbf{Center point.}
For low, flat robots whose bounding box is effectively their footprint, the anchor is the box center $\bigl(b_x W_{\mathrm{img}},\; b_y H_{\mathrm{img}}\bigr)$, with $W_{\mathrm{img}}$ and $H_{\mathrm{img}}$ the image width and height in pixels.

\noindent\textbf{Top--bottom.}
For tall or self-occluding objects the anchor depends on where the object sits in the frame. A box in the upper image half is viewed near-horizontally, and its bottom edge is a trustworthy ground contact, so the anchor is the bottom-center $\bigl(b_x W_{\mathrm{img}},\; (b_y + b_h/2) H_{\mathrm{img}}\bigr)$. A box in the lower image half is close to the camera under a steep view, where the base is routinely occluded or clipped while the top of the object stays crisp; there we anchor on the reliable top edge and descend by the object's projected height. Let $\tilde{\mathbf{X}}_h = (X, Y, h, 1)^{\!\top}$ be the world point at height $h$ above the ground point $(X,Y)$, and let $\mathbf{p}^{(2)}_c$ and $\mathbf{p}^{(3)}_c$ denote the second and third rows of $\mathbf{P}_c$. The image row of that point, and the apparent pixel height of a vertical segment of length $h$, are
\vspace{-0.3cm}
\begin{equation}
  v(h) \;=\; \frac{\mathbf{p}^{(2)}_c \tilde{\mathbf{X}}_h}{\mathbf{p}^{(3)}_c \tilde{\mathbf{X}}_h},
  \qquad \Delta v_c(X,Y,h) \;=\; \bigl|\,v(0) - v(h)\,\bigr|,
  \label{eq:apparent_height}
  \vspace{-0.3cm}
\end{equation}
where $v(h)$ is the image row of the top of that segment standing at $(X,Y)$ and $\Delta v_c$ its apparent pixel height. Seeding the footprint at the bottom-center back-projection and taking $h = H_k$, the constant world height prior of class $k$ (defined with the other class extents in~\cref{subsection:height_and_yaw_angle_estimation}), the anchor row becomes
\vspace{-0.3cm}
\begin{equation}
  v_{\text{anchor}} \;=\; \max\Bigl(\,(b_y - b_h/2)\,H_{\mathrm{img}} \;+\; \Delta v_c\bigl(\pi_c(\mathbf{u}_{\text{bot}}),\, H_k\bigr),\;\; (b_y + b_h/2)\,H_{\mathrm{img}} \,\Bigr),
  \label{eq:anchor_row}
  \vspace{-0.3cm}
\end{equation}
where $\mathbf{u}_{\text{bot}}$ is the box bottom-center pixel and $H_k$ the class height prior of~\cref{eq:lift} in~\cref{subsection:height_and_yaw_angle_estimation}. The $\max$ enforces that the anchor never rises above the visible box bottom, while permitting it to fall below that edge, exactly the intended correction when the true base is clipped out of the box. Every anchor pixel is back-projected with $\mathbf{H}_c$ through~\cref{eq:backproj} and discarded if it is degenerate or falls outside the camera's ground-coverage zone.
This third strategy is powerful but circular: it derives the ground contact \emph{from} the class height prior. The estimator in~\cref{subsection:height_and_yaw_angle_estimation} breaks that circularity for persons by estimating height from a ground point constructed independently of it.

\subsubsection{3.4.2 Cross-View Association on the Ground Plane}

At each frame, projected observations $o_i=(c_i,k_i,\ell_i,\mathbf{x}_i,s_i,\mathbf{F}_i,\boldsymbol{\nu}_i)$, encoding camera, class, local identity, ground position, confidence, descriptor, and part visibility, are associated within each class before temporal tracking. Appearance is compared by a visibility-weighted cosine distance that ignores parts occluded in either view:
\vspace{-0.4cm}
\begin{equation}
  d_{\text{app}}(o_a, o_b) \;=\; \left.\sum_{p=1}^{6} \nu_{a,p}\,\nu_{b,p}\,\bigl(1 - \langle \mathbf{f}_{a,p},\, \mathbf{f}_{b,p}\rangle\bigr) \;\middle/\; \sum_{p=1}^{6} \nu_{a,p}\,\nu_{b,p}\right.,
  \label{eq:app_dist}
  \vspace{-0.4cm}
\end{equation}
where $p$ indexes six body parts; since the descriptor rows are unit-norm, $d_{\text{app}}\in[0,2]$, while no co-visible part gives $+\infty$. Observations from different cameras are admissible when $\|\mathbf{x}_i-\mathbf{x}_j\|_2<\theta^{\text{bev}}_k$, a per-class BEV merge radius, and, for classes in $\mathcal{K}_{\text{reid}}$ (the classes with part-based appearance descriptors), $d_{\text{app}}(o_i,o_j)<\theta_{\text{grp}}$, an appearance grouping gate. Single-linkage clustering~\cite{sibson1973slink} over a union--find structure~\cite{tarjan1975efficiency} processes pairs by increasing ground distance and rejects merges between clusters already containing the same camera.
Each cluster $\mathcal{G}_\omega$ becomes $\omega=(k_\omega,\mathbf{x}_\omega,\mathbf{F}_\omega,\boldsymbol{\nu}_\omega,\mathcal{G}_\omega)$, with confidence-weighted position $\mathbf{x}_\omega=\frac{\sum_{m\in\mathcal{G}_\omega}s_m\mathbf{x}_m}{\sum_{m\in\mathcal{G}_\omega}s_m}$ (or an unweighted mean if the denominator is zero). Part descriptors are fused as $\mathbf{f}_{\omega,p}=\frac{\sum_{m\in\mathcal{G}_\omega}\nu_{m,p}\mathbf{f}_{m,p}}{\|\sum_{m\in\mathcal{G}_\omega}\nu_{m,p}\mathbf{f}_{m,p}\|_2}$ with $\nu_{\omega,p}=\max_{m\in\mathcal{G}_\omega}\nu_{m,p}$; this combines complementary views without imputing occluded parts, and when no member sees part $p$ the fused weight is $\nu_{\omega,p}=0$, so the part is ignored by~\cref{eq:app_dist}.

\subsubsection{3.4.3 Global Multi-View Tracking}

A global track $\tau$ holds a class $k_\tau$, a ground position $\mathbf{x}_\tau$, a velocity $\mathbf{v}_\tau$ ($\text{m}\,\text{frame}^{-1}$), a member map $\mathcal{M}_\tau : c \mapsto \ell$ of the local tracklets feeding it, an appearance gallery $\mathcal{A}_\tau$ of up to $N_{\text{gal}}$ recent descriptor--visibility pairs, and a missing age $a_\tau$ of consecutive unmatched frames. Each frame's world observations bind to tracks in two passes.

\noindent\textbf{Pass A: carry-forward by local identity.}
Per observation $\omega$ and same-class track $\tau$ we count $\omega$'s members $(c,\ell)$ already in $\mathcal{M}_\tau$ and bind pairs greedily in descending count, each observation and track consumed once. This pass is the primary defense against identity switches: a track survives while \emph{any} one of its cameras holds it.

\noindent\textbf{Pass B: gated assignment for the remainder.}
Whatever Pass~A leaves is resolved per class. The prediction $\hat{\mathbf{x}}_\tau = \mathbf{x}_\tau + \mathbf{v}_\tau$ gives $d_{\text{bev}}(\omega,\tau) = \|\mathbf{x}_\omega - \hat{\mathbf{x}}_\tau\|_2$, gated by $g(\tau) = \theta^{\text{bev}}_{k} + \frac{V_{\max}}{f_{\mathrm{r}}}a_\tau$, exactly the distance coverable while unobserved at maximum plausible speed $V_{\max}$ ($\text{m}\,\text{s}^{-1}$) and frame rate $f_{\mathrm{r}}$, so a track survives long occlusions without an indiscriminate gate. Appearance uses the nearest gallery distance $d_{\text{gal}}(\omega,\tau) = \allowbreak \min_{(\mathbf{F}',\boldsymbol{\nu}') \in \mathcal{A}_\tau} \allowbreak d_{\text{app}}\bigl((\mathbf{F}_\omega,\boldsymbol{\nu}_\omega),\,\allowbreak(\mathbf{F}',\boldsymbol{\nu}')\bigr)$, where $d_{\text{app}}$ of~\cref{eq:app_dist} depends on its arguments only through the descriptor--visibility pair and is $+\infty$ absent a gallery or co-visible part; the minimum, not the mean, lets one confident past view re-identify across pose or illumination change. Pairs are gated out by $d_{\text{bev}} > g(\tau)$, or by $k \in \mathcal{K}_{\text{reid}} \wedge \theta_{\text{reid}} < d_{\text{gal}} < \infty$ for appearance gate $\theta_{\text{reid}}$. Survivors cost $\mathbf{C}[\omega,\tau] = \beta\frac{d_{\text{bev}}(\omega,\tau)}{g(\tau)} + \allowbreak (1-\beta)\bar{d}_{\text{gal}}(\omega,\tau)$ for $k \in \mathcal{K}_{\text{reid}}$ and $\mathbf{C}[\omega,\tau] = d_{\text{bev}}(\omega,\tau)$ otherwise, with $\bar{d}_{\text{gal}} = d_{\text{gal}}$ when finite and a neutral $1$ otherwise. Normalizing by $g(\tau)$ puts geometry on a bounded scale comparable to the appearance term's $[0,2]$ range, so a single weight $\beta \in [0,1]$ balances the two across classes with very different gate radii. The Hungarian algorithm~\cite{kuhn1955hungarian} minimizes $\sum \mathbf{C}$ exactly; $\infty$-sentinel pairs are discarded after the assignment is solved.

\noindent\textbf{Trajectory Lifecycle Dynamics.}
An unmatched observation spawns a track only when $\min_{\tau:\,k_\tau = k_\omega}\|\mathbf{x}_\omega - \mathbf{x}_\tau\|_2 \geq \delta_{\text{new}}$, the minimum birth separation (m); this blocks the rebirth of a momentarily unmatched track beside itself under a fresh identity. An unmatched track instead dead-reckons, $\mathbf{x}_\tau \leftarrow \mathbf{x}_\tau + \mathbf{v}_\tau$, $a_\tau \leftarrow a_\tau + 1$, and retires once $a_\tau > A_{\max}$. A matched track takes $\mathbf{x}^{+}_\tau = \alpha\mathbf{x}_\omega + (1-\alpha)\mathbf{x}_\tau$ for smoothing factor $\alpha \in (0,1]$ and $\mathbf{v}_\tau = \allowbreak \frac{\mathbf{x}^{+}_\tau - \mathbf{x}_\tau}{\max(1,\; t - t^{\text{last}}_\tau)}$ from its previous sighting frame $t^{\text{last}}_\tau$, then $\mathbf{x}_\tau \leftarrow \mathbf{x}^{+}_\tau$; its members refresh $\mathcal{M}_\tau$, its fused descriptor enters $\mathcal{A}_\tau$, and $a_\tau$ resets to zero.

\noindent\textbf{A closed-world cardinality prior.}
In a controlled warehouse the per-class count $N_k$ is often known a priori: trivially in simulation, by inspection for a staged capture. A fixed-cardinality variant then constrains each class's live track set $\mathcal{T}_k$,
\vspace{-0.2cm}
\begin{equation}
  |\mathcal{T}_k| \;\leq\; N_k \quad \forall k, \qquad N_k = \infty \;\text{ if } k \text{ is uncapped},
  \label{eq:cardinality}
  \vspace{-0.2cm}
\end{equation}
under which capped tracks coast rather than retire, so the count converges to exactly $N_k$; contested slots go to the most confident observations, and a stale slot is reclaimed past a missing-age threshold. Association thus becomes assigning a \emph{fixed} identity set to the frame's evidence rather than managing tracks from detections. This regime is sharpest with the Pass-B gates dropped: the solve force-assigns every track to its best observation, and a track coasts only when the frame yields fewer observations than its class has tracks. The prior is exact in simulation, where the census is a fact of the scene description, and precisely what a real deployment cannot assume; the real scene therefore runs the adaptive tracker with all gates active.

\noindent\textbf{Finalization.}
In an offline post-processing pass after the last frame (the only acausal step, applied once per sequence before submission), each track's sightings become a per-frame position map; gaps under $G_{\max}$ frames are linearly interpolated between bracketing sightings, and longer ones are left unfilled instead of being bridged through space the object may not have occupied.
\begin{table}[t]
  \centering
  \caption{The synthetic-to-real configuration surface: parameter settings used in the synthetic and real regimes.}
  \label{tab:sim2real_surface}
  \vspace{-0.4cm}
  \small
  \begin{tabular}{@{}lll@{}}
    \toprule
    Parameter & Synthetic regime & Real regime \\
    \midrule
    Cameras per scene & $10$--$20$ & $4$--$7$ \\
    Intrinsics & ideal pinhole, cloned & per-camera, $f_x \neq f_y$ \\
    Person ground anchor & bbox $+$ height pole & ankle skeleton \\
    Ankle confidence $\theta_{\text{ank}}$ & $0.5$ & $0.8$ \\
    Skeleton gate $n_{\text{kp}}$ & $3$ & $8$ \\
    Position smoothing $\alpha$ & $1.0$ (none) & $0.6$ \\
    Appearance gates $\theta_{\text{grp}},\,\theta_{\text{reid}}$ & $>2$ (inactive) & $0.8$,\ $1.6$ (active) \\
    Object cardinality $N_k$ & known and enforced & unknown; adaptive \\
    Gap interpolation $G_{\max}$ & $0$ frames & $60$ frames \\
    \bottomrule
  \end{tabular}
  \vspace{-0.6cm}
\end{table}

\subsubsection{3.4.4 The Synthetic-to-Real Configuration Surface}
The domain gap shifts the relative reliability of geometry and appearance (\cref{tab:sim2real_surface}): synthetic scenes (exact calibration, broad overlap, known counts) let geometry and the closed-world prior suffice, while real scenes (estimated calibration, sparse overlap, unknown counts) need smoothing and active appearance gating (toggled by the cosine threshold of~\cref{eq:app_dist}). The pipeline thus adapts by shifting trust between geometry and appearance, not by retraining the feature extractor.

\subsection{Height and Yaw Angle Estimation}
\label{subsection:height_and_yaw_angle_estimation}

Each track position $(X,Y)$ at frame $t$ is lifted to an axis-parameterized 3D box $(X, Y, Z, w, l, h, \psi)$. For yaw $\psi$, we fit a second-degree polynomial to the previous 45 track positions, differentiate at the current point, and take the arctangent. The baseline lift assigns constant class extents,
\vspace{-0.3cm}
\begin{equation}
  (w, l, h) \;=\; (W_k,\, L_k,\, H_k), \qquad Z \;=\; \tfrac{h}{2},
  \label{eq:lift}
\vspace{-0.2cm}
\end{equation}
where $(W_k, L_k, H_k)$ are class $k$'s width, length, and height priors, the per-class medians of the previous year's warehouse ground-truth 3D box scales (the one class lacking a counterpart there is given a manual estimate), and $Z=h/2$ places a floor-standing object's centroid at half its height. Imported wholesale, this synthetic prior is the second gap point: the shape prior.

\noindent\textbf{Monocular metric person height.}
For persons this prior becomes a per-frame calibration estimate. To break the circularity of the $H_k$-dependent anchor in~\cref{eq:anchor_row}, we form an independent ground point $\tilde{\mathbf{X}}_0=(X,Y,0,1)^{\!\top}$ from the ankle midpoint; with $a_v=\mathbf{p}^{(2)}_c\tilde{\mathbf{X}}_0$, $b_v=(\mathbf{P}_c)_{23}$, $\gamma_v=\mathbf{p}^{(3)}_c\tilde{\mathbf{X}}_0$, and $d_v=(\mathbf{P}_c)_{33}$, $v(h)=(a_v+b_v h)/(\gamma_v+d_v h)$, and matching the detection's top row $v_{\text{top}}$ yields
\vspace{-0.3cm}
\begin{equation}
  \hat{h} \;=\; \frac{v_{\text{top}}\,\gamma_v \;-\; a_v}{b_v \;-\; v_{\text{top}}\, d_v},
  \label{eq:height}
  \vspace{-0.3cm}
\end{equation}
which inverts~\cref{eq:apparent_height} without learned depth or 3D ground truth. A sample is accepted only if the detection is untruncated, its confidence exceeds the track-confidence threshold $\theta_{\text{trk}}$, both ankles exceed $\theta_{\text{ank}}$, a head keypoint exceeds $\theta_{\text{pose}}$, the back-projection is valid and in-zone, and $\hat{h}\in[h_{\min},h_{\max}]$; failing samples are rejected, not clamped, which leaves a small positive bias from ankle elevation. Track $\tau$ takes the cross-camera median $\hat{h}_{\tau,t}=\operatorname{median}\{\hat{h}_m\mid m\in\mathcal{G}_{\omega},\ \hat{h}_m\ \text{accepted}\}$ over its members $\mathcal{G}_{\omega}$, else $H_k$ with no cross-frame propagation. The box then scales as $(w,l,h,Z)=\bigl(\tfrac{W_k}{H_k}\hat{h}_{\tau,t},\,\tfrac{L_k}{H_k}\hat{h}_{\tau,t},\,\hat{h}_{\tau,t},\,\tfrac{\hat{h}_{\tau,t}}{2}\bigr)$, which preserves the class aspect ratio.
\begin{figure}[t]
    \centering
    \begin{minipage}{0.8\linewidth}
        \centering
        \includegraphics[width=\linewidth,valign=c]{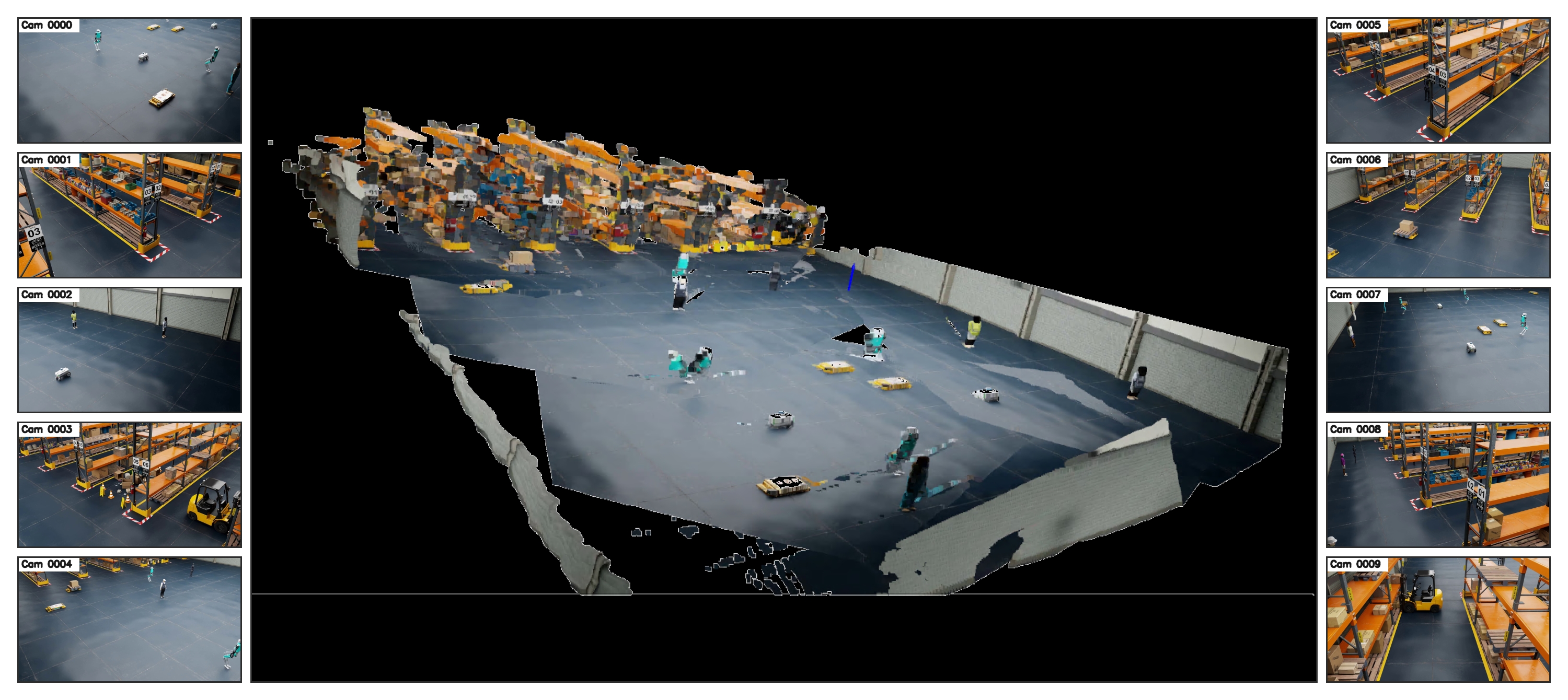}\\[-1mm]
        \small (a) Synthetic scene
    \end{minipage}
    \begin{minipage}{0.8\linewidth}
        \centering
        \includegraphics[width=\linewidth,valign=c]{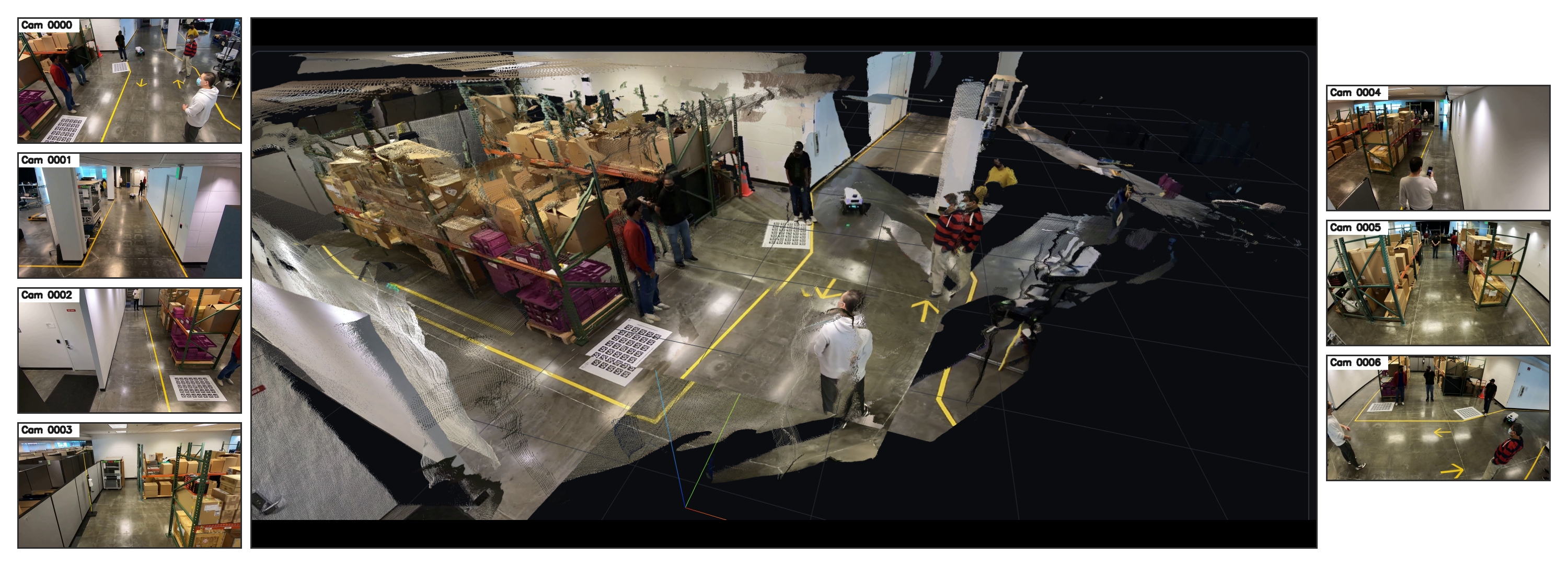}\\[-1mm]
        \small (b) Real-world scene
    \end{minipage}
	\vspace{-0.3cm}
    \caption{Multi-view RGB inputs and DA3-estimated point clouds for (a) the synthetic scene Warehouse~025 and (b) the real-world scene Warehouse~027 at frame 0.}
    \label{fig:da3_multiview_pointcloud}
	\vspace{-0.6cm}
\end{figure}
\subsection{Point-Cloud-Guided 3D Box Refinement}
\label{subsection:point_cloud_guided_3d_box_refinement}

After multi-view tracking, each object has a stable identity and size but a coarse center, obtained by projecting a 2D image point to the floor via intrinsics $\mathbf{K}_c$ and extrinsics $\mathbf{E}_c$. This projection fails when the point is not a true ground contact (elevated Transporters, occluded feet, distortion, calibration residuals), producing planar position and yaw errors that lower 3D IoU despite correct association. Since Track~1 withholds depth at inference~\cite{Tang26AICity26}, we refine footprint and yaw from an RGB-estimated point cloud, preserving identity and size; \cref{fig:point_cloud_guided_refinement_examples} shows typical pre-refinement misalignments.

\noindent\textbf{DA3-Based Metric Point Cloud Construction.}
We run Depth Anything~3 (DA3)~\cite{lin2026depth} in a pose-conditioned multi-view setting with the Nested Giant-Large~1.1 checkpoint, which couples any-view geometry with metric-depth scaling. Per synchronized frame, all camera images are processed with the provided $\mathbf{K}_c$ and $\mathbf{E}_c$, and predicted depths are back-projected into a common world frame and fused into a colored point cloud (\cref{fig:da3_multiview_pointcloud}; details in~\cref{sec:point_cloud_construction_details}).

\begin{figure}[t]
    \centering
    \includegraphics[width=\textwidth,valign=c]{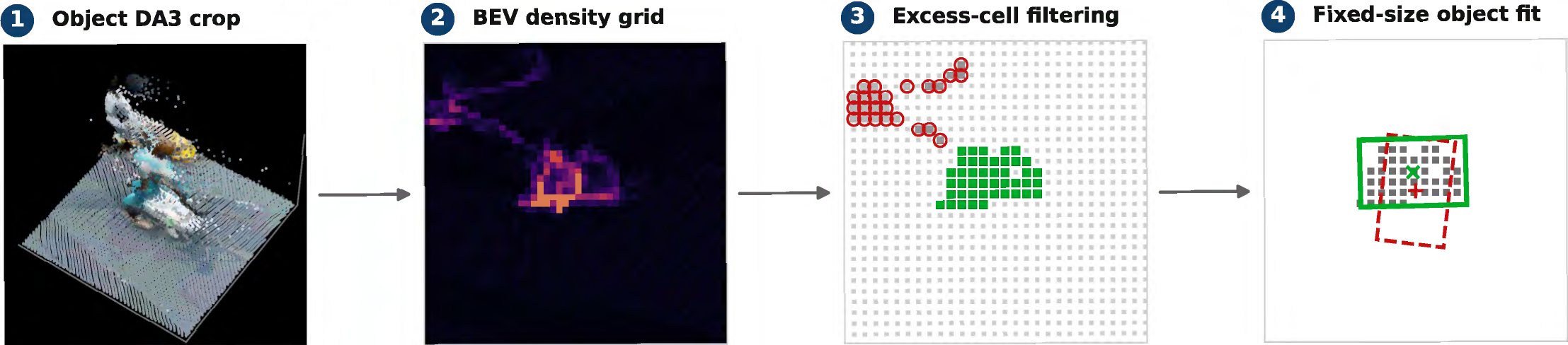}
    \caption{Local fixed-shape object refinement pipeline for the synthetic scene. Starting from a colored DA3 \texttt{.ply} crop around a tracked object, we project the crop to a BEV density grid, keep density-excess cells after floor and ghost filtering, and fit a fixed-size box to the remaining support. The red dashed box is the multi-view tracking prior, and the green box is the refined result.}
    \label{fig:local_refinement_pipeline}
	\vspace{-0.4cm}
\end{figure}

\noindent\textbf{Synthetic Scene Refinement.}
Reliable calibration and controlled categories keep the class-size priors of~\cref{eq:lift} valid, so DA3 refines only footprint and yaw. For fixed-shape objects, we crop points around the tracked box, project them to a BEV density grid, drop floor, background, and weak components, and fit the prior-sized footprint near the tracked pose, reverting to the tracker output when support is weak (\cref{fig:local_refinement_pipeline}). For dynamic-shape objects (persons, humanoid robots), the tracker center is steadier than the sparse articulated cloud, so DA3 corrects only standing yaw or one-frame jumps and keeps the prior (\cref{fig:point_cloud_guided_refinement_examples}(a)).

\noindent\textbf{Real-World Scene Refinement.}
Moving targets are mainly persons, whose centers are sensitive to distorted rays, calibration residuals, and occluded feet; parked vehicles keep fixed hand-verified poses. After removing static background via temporal depth statistics, each person track's BEV window is resolved to its densest mode by seeded coarse-to-fine mean-shift~\cite{comaniciu_mean_2002}: the tracker seeds the first frame, later frames reuse the refined state, and a causal temporal state bridges short losses and suppresses jitter. DA3 corrects the center while keeping the class-size prior (\cref{fig:point_cloud_guided_refinement_examples}(b)).

\subsection{Filtering Predictions Based on the BEV Map}
\label{subsection:filter_predictions_based_on_bev_map}

\begin{figure}[t]
	\begin{center}
		\resizebox{0.9\linewidth}{!}{
			\begin{tabular}{cc}
				\includegraphics[width=0.4\linewidth,valign=c]{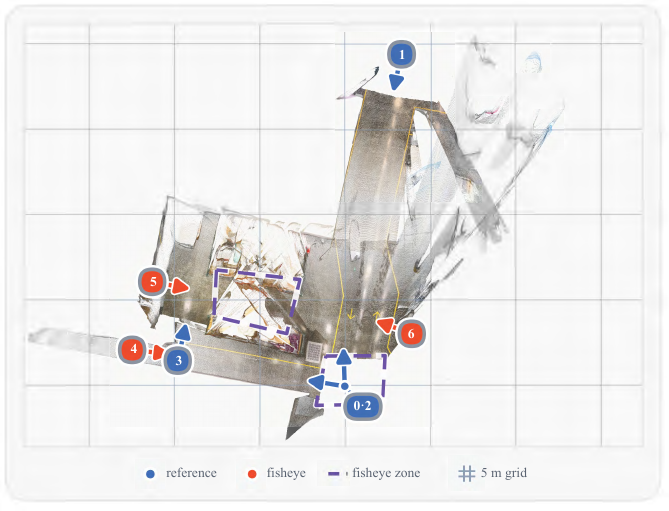} & 
				\includegraphics[width=0.4\linewidth,valign=c]{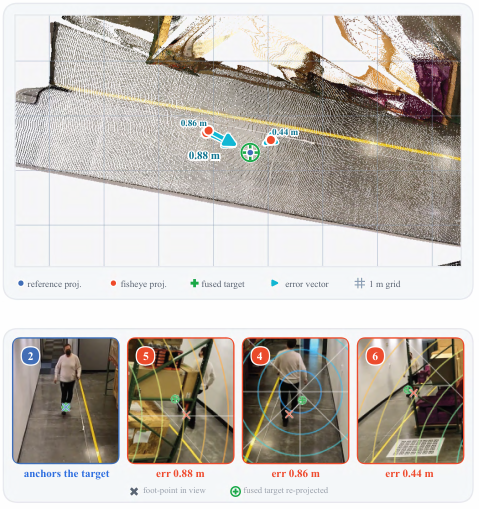}   \\
				(a) BEV floor map: cameras and fisheye zones &
				(b) BEV consistency: one matched object
			\end{tabular}
		}
		\vspace{-0.2cm}
		\caption{BEV use of camera grouping. (a)~Floor map of reference (blue) and fisheye (red) cameras; dashed violet marks fisheye-only zones, where objects retain their fisheye BEV positions. (b)~For a matched object, the reference view anchors the fused target, and each fisheye projection's error is measured in ground-plane meters.}
	\label{fig:distortion_bev}
	\end{center}
	\vspace{-1.0cm}
\end{figure}

A per-class cap (\cref{eq:cardinality}) or weak detections can emit a box no camera genuinely supports: a phantom that inflates false positives. We test every emitted box against the cameras whose ground-plane coverage includes it, via per-camera visibility maps on the BEV floor map. The filter runs per frame on the emitted box set only and never alters a box's identity, size, or pose; it proceeds in three steps:

\noindent\textbf{Definition of Camera Visibility Maps}:
For each camera $c$ we record the ground region it observes: its projected field of view intersected with the hand-annotated coverage polygons of the BEV floor map (\cref{fig:distortion_bev}), which also mark the fisheye-only zones. Rasterizing over all cameras gives, per BEV cell, the set of cameras that see it: a visibility map $\mathcal{V}(\mathbf{x})$ indexed by ground position $\mathbf{x}$.

\noindent\textbf{Object Verification and Filtering}:
Each emitted box at $\mathbf{x}_\tau$ gets covering cameras $\mathcal{C}(\tau,t)=\mathcal{V}(\mathbf{x}_\tau)$; when this set is empty, all cameras the box geometrically projects onto are used instead. A person box is re-projected into these views and kept only if some re-extracted crop shows a visible ankle,
\vspace{-0.3cm}
\begin{equation}
  \operatorname{keep}(\tau,t) \iff \mathcal{C}(\tau,t)\neq\varnothing
  \;\land\; \max_{c \,\in\, \mathcal{C}(\tau,t)} \max_{j \,\in\, \{15,16\}} s_{c,j} > \theta_{\text{vis}}
  \quad (k_\tau\ \text{a person class}),
  \label{eq:keep}
  \vspace{-0.5cm}
\end{equation}
where $s_{c,j}$ is the score of ankle keypoint $j$ in camera $c$ and $\theta_{\text{vis}}$ the visibility threshold. Unconfirmed person boxes are removed, and boxes of all other classes pass unchanged; this causal per-frame check suppresses forced-assignment phantoms while leaving every surviving box untouched.

\noindent\textbf{Fisheye BEV Consistency} (\cref{fig:distortion_bev}):
Some ground regions are seen only by fisheye cameras and have no reference anchor; these blind spots are manually annotated as BEV polygons (dashed violet in~\cref{fig:distortion_bev}(a)), and objects inside keep their fisheye-detection BEV position instead of being refined toward an unsupported target. Where both groups overlap (\cref{fig:distortion_bev}(b)), matched observations are projected to BEV and the fisheye-versus-reference disagreement is measured in meters, the error BEV tracking consumes; a foot point is then corrected by undistorting it before floor projection, refining only the distortion while intrinsics and extrinsics stay fixed.
\section{Experiments \& Discussion}
\label{section:experiments_discussion}


\subsection{Dataset and Evaluation Metrics}
\label{subsection:datasets}

MTMC Tracking 2026~\cite{Tang26AICity26} provides synchronized 1080p/30-fps RGB videos from calibrated cameras, top-down maps, and 2D/3D annotations. Under its RGB-only protocol, we evaluate Syn2RealTrack on expanded synthetic and hidden real scenes using HOTA~\cite{luiten_hota_2021}, DetA, AssA, and LocA (higher is better).

\begin{table}[t]
	\caption{Leaderboard of Multi-Camera 3D Perception, AI City Challenge 2026 Track~1, ranked by 3D HOTA (\%). Our entry (shaded, bold) ranks second.}
	\label{tab:final_score_leaderboard}
	\vspace{-0.6cm}
	\begin{center}
		\resizebox{0.7\columnwidth}{!}{%
		\begin{tabular}{@{}ccccccc@{}}
			\toprule
			\textbf{Rank} & \textbf{ID} & \textbf{Name} & \textbf{3D HOTA (\%)} & \textbf{DetA (\%)} & \textbf{AssA (\%)} & \textbf{LocA (\%)} \\
      \midrule
      1 & 289 & EVA & 56.5447 & 55.6444 & 49.3929 & 79.5558 \\
			\rowcolor[HTML]{EFEFEF} 2 & 34 & SKKU-AL-T1 & \textbf{52.0118} & 45.3056 & 56.5047 & 76.2410 \\
			3 & 130 & Playbox & 38.0105 & 40.2592 & 31.0978 & 75.1778 \\
			4 & 4 & QDTers & 34.1845 & 29.4663 & 33.1122 & 17.6841 \\
			5 & 149 & Calix & 33.7654 & 30.8748 & 30.9816 & 50.4720 \\
      \bottomrule
		\end{tabular}%
	}
	\end{center}
	\vspace{-0.7cm}
\end{table}
\subsection{Quantitative Results}
\label{subsection:quantitative_result}
\cref{tab:final_score_leaderboard} reports the official Track~1 leaderboard, ranked by 3D HOTA. Our team, SKKU-AL-T1, ranks second at 52.0118\%, trailing first-place EVA by 4.53 points but leading third place by 14.00. The field thus splits into two tiers: our gap to the rest is roughly three times our gap to the top.
\begin{table}[t]
	\begin{minipage}[t]{0.48\columnwidth}
		\caption{Impact of the detection backbone and training resolution on the Track~1 evaluation server. The shaded row is the setting adopted in~\cref{subsection:2d_object_detection}.}
		\label{tab:ablation_detection}
		\vspace{-0.8cm}
		\begin{center}
			\resizebox{\linewidth}{!}{%
			\begin{tabular}{@{}lccc@{}}
				\toprule
				\textbf{Model} & \textbf{Train Size} & \textbf{Infer Size} & \textbf{HOTA (\%)} \\
				\midrule
				\multirow{2}{*}{YOLOv26} & 1600 & 1920 & 50.5285 \\
				                         & 1920 & 1920 & 50.7207 \\
				\midrule
				\multirow{2}{*}{RF-DETR} & 1600 & 1920 & 50.5913 \\
				 & 1920 & 1920 & \textbf{50.8078} \\
				\bottomrule
			\end{tabular}%
		}
		\end{center}
	\end{minipage}%
	\hfill%
	\begin{minipage}[t]{0.48\columnwidth}
		\caption{Impact of the association cues used for single-view tracking on the Track~1 evaluation server. The shaded row is the setting adopted in~\cref{subsection:single_view_tracking}.}
		\label{tab:ablation_association}
		\vspace{-0.6cm}
		\begin{center}
			\resizebox{0.7\linewidth}{!}{%
			\begin{tabular}{@{}ccccc@{}}
				\toprule
				\textbf{Re-ID} & \textbf{Pose} & \textbf{gIoU} & \textbf{dIoU} & \textbf{HOTA (\%)} \\
				\midrule
				\textcolor{red}{\ding{55}} & \textcolor{red}{\ding{55}} & \textcolor{red}{\ding{55}} & \textcolor{red}{\ding{55}} & 50.8078 \\
				\textcolor{red}{\ding{55}} & \textcolor{red}{\ding{55}} & \textcolor{teal}{\ding{51}} & \textcolor{red}{\ding{55}} & 51.1145 \\
				\textcolor{red}{\ding{55}} & \textcolor{red}{\ding{55}} & \textcolor{red}{\ding{55}} & \textcolor{teal}{\ding{51}} & 50.9577 \\
				\textcolor{red}{\ding{55}} & \textcolor{red}{\ding{55}} & \textcolor{teal}{\ding{51}} & \textcolor{teal}{\ding{51}} & 50.8204 \\
				\textcolor{teal}{\ding{51}} & \textcolor{red}{\ding{55}} & \textcolor{teal}{\ding{51}} & \textcolor{teal}{\ding{51}} & 51.2906 \\
				\textcolor{red}{\ding{55}} & \textcolor{teal}{\ding{51}} & \textcolor{teal}{\ding{51}} & \textcolor{teal}{\ding{51}} & 51.3205 \\
				\rowcolor[HTML]{EFEFEF} \textcolor{teal}{\ding{51}} & \textcolor{teal}{\ding{51}} & \textcolor{teal}{\ding{51}} & \textcolor{teal}{\ding{51}} & \textbf{51.8568} \\
				\bottomrule
			\end{tabular}%
		}
		\end{center}
	\end{minipage}
	\vspace{-0.6cm}
\end{table}
\begin{table}[t]
	\begin{minipage}[t]{0.48\columnwidth}
		\caption{Impact of monocular metric person height estimation (\cref{eq:height}); the shaded row is the setting adopted in~\cref{subsection:height_and_yaw_angle_estimation}.}
		\label{tab:ablation_height}
		\vspace{-0.8cm}
		\begin{center}
			\resizebox{\linewidth}{!}{%
			\begin{tabular}{@{}lc@{}}
				\toprule
				\textbf{Person height source} & \textbf{HOTA (\%)} \\
				\midrule
				Constant class prior (\cref{eq:lift}) & 51.8568 \\
				$+$ monocular estimate, Warehouse~023 \& 024 & 51.9546 \\
				\rowcolor[HTML]{EFEFEF} $+$ monocular estimate, Warehouse~025 & \textbf{51.9898} \\
				\bottomrule
			\end{tabular}%
		}
		\end{center}
	\end{minipage}%
	\hfill%
	\begin{minipage}[t]{0.48\columnwidth}
		\caption{Impact of BEV-map-based prediction filtering (\cref{subsection:filter_predictions_based_on_bev_map}).}
		\label{tab:ablation_bev_filter}
		\vspace{-0.8cm}
		\begin{center}
			\resizebox{\linewidth}{!}{%
			\begin{tabular}{@{}lc@{}}
				\toprule
				\textbf{Filtering stage} & \textbf{HOTA (\%)} \\
				\midrule
				No filtering & 51.8813 \\
				$+$ visibility-zone \& static-object filtering & 52.0100 \\
				\rowcolor[HTML]{EFEFEF} $+$ 3D-box coverage verification & \textbf{52.0118} \\
				\bottomrule
			\end{tabular}%
		}
		\end{center}
	\end{minipage}
	\vspace{-0.4cm}
\end{table}
\subsection{Ablation Study}
\label{subsection:ablation_study}

Since hidden-test annotations are unavailable, each ablation is a single evaluation-server submission, comparable with~\cref{tab:final_score_leaderboard}. Sequential configurations with minor scene-specific tuning give stage-wise, not independent, effects; we keep four-decimal precision to expose small differences.

\noindent\textbf{2D Detection Backbone and Input Resolution.}
This experiment selects the detection backbone and training resolution. At an inference resolution of $1920$, \cref{tab:ablation_detection} shows RF-DETR beating YOLOv26 at both training resolutions; matching the training to the inference resolution adds $0.19$ and $0.22$ HOTA, motivating the $1920$/$1920$ setting (\cref{subsection:2d_object_detection}). The backbones differ by only $0.09$ HOTA, so the detector choice is not decisive.

\noindent\textbf{Single-View Association Cues.}
\cref{tab:ablation_association} isolates the single-view affinity terms of~\cref{subsection:single_view_tracking}. Over the $50.8078$ baseline, generalized IoU (gIoU)~\cite{rezatofighi2019giou} and distance IoU (dIoU)~\cite{zheng2020diou} add $+0.31$ and $+0.15$ HOTA alone, but combined without appearance reach only $50.8204$. Re-identification or pose guidance raises HOTA to $51.2906$ and $51.3205$; both together give the best $51.8568$ ($+1.05$ over baseline), the final configuration.

\noindent\textbf{Monocular Person Height Estimation.}
\cref{tab:ablation_height} replaces the constant prior in~\cref{eq:lift} with the per-frame estimate of~\cref{eq:height}, enabled scene by scene. HOTA improves monotonically, by $+0.13$ in total over the constant-prior baseline: calibration alone beats the synthetic prior without depth or supervision. The estimator adjusts box height and centroid, not identity, so association is essentially unaffected. A residual scene-dependent ground-plane error remains, which the point-cloud refinement of~\cref{subsection:point_cloud_guided_3d_box_refinement} targets.

\noindent\textbf{BEV-Map Prediction Filtering.}
\cref{tab:ablation_bev_filter} evaluates the false-positive filter of~\cref{subsection:filter_predictions_based_on_bev_map}, which removes boxes induced by the closed-world prior of~\cref{eq:cardinality} at camera-unsupported locations. Disabling it costs $0.13$ HOTA; in the cumulative decomposition of~\cref{tab:ablation_cumulative}, the same stage adds $+0.23$ DetA while lowering AssA by only $0.01$. This asymmetry supports the causal, per-frame test of~\cref{eq:keep}, which suppresses phantoms without terminating identities held elsewhere. The final coverage check adds $+0.0018$ HOTA at negligible overhead.
\begin{table}[t]
	\caption{Stage-wise ablation of the full pipeline on the Track~1 evaluation server. Each row adds one component to the row above it; the shaded row~6 is the submitted system of~\cref{tab:final_score_leaderboard}.}
	\label{tab:ablation_cumulative}
	\vspace{-0.9cm}
	\begin{center}
		\resizebox{\columnwidth}{!}{%
		\begin{tabular}{@{}clccccc@{}}
			\toprule
			\textbf{\#} & \textbf{Configuration} & \textbf{HOTA (\%)} & \textbf{DetA (\%)} & \textbf{AssA (\%)} & \textbf{LocA (\%)} & \textbf{$\Delta$ HOTA} \\
			\midrule
			1 & RF-DETR detector only                  & 50.8078 & 42.4111 & 56.2599 & 74.1915 & --      \\
			2 & $+$ gIoU association                   & 51.1145 & 43.6031 & 55.8617 & 75.0459 & $+0.31$ \\
			3 & $+$ all association cues               & 51.8568 & 44.7921 & 56.4608 & \textbf{76.2410} & $+0.74$ \\
			4 & $+$ cross-view ground-plane clustering & 51.9205 & 45.1611 & 56.4601 & 76.2053 & $+0.06$ \\
			5 & $+$ monocular person height estimation & 51.9898 & 45.0712 & \textbf{56.5190} & 76.2397 & $+0.07$ \\
			\rowcolor[HTML]{EFEFEF} 6 & $+$ BEV-map prediction filtering & \textbf{52.0118} & \textbf{45.3056} & 56.5047 & \textbf{76.2410} & $+0.02$ \\
			\bottomrule
		\end{tabular}%
	}
	\end{center}
	\vspace{-1.2cm}
\end{table}

\noindent\textbf{Contribution of Each Pipeline Stage.}
The cumulative ablation in~\cref{tab:ablation_cumulative} raises HOTA from $50.8078$ to $52.0118$ ($+1.20$), of which single-view association supplies $+1.05$ ($87.1\%$). The remaining stages add smaller, detection-driven gains: cross-view ground-plane clustering $+0.06$, monocular person height $+0.07$, and BEV-map filtering $+0.02$, the last through a $+0.23$ DetA gain. Across all stages DetA rises from $42.41$ to $45.31$ ($+2.90$), while AssA stays within $56.46$--$56.52$ once all association cues are enabled.

\section{Conclusion}
\label{section:conclusion}
Syn2RealTrack addresses synthetic-to-real gaps in calibration, shape priors, and object census, achieving $52.0118\%$ 3D HOTA and second place in AI City Challenge 2026 Track~1. It remains sensitive to calibration and RGB-derived depth, severe occlusion, frequent entry/exit, and closed-world census assumptions; future work will target uncertainty-aware calibration and depth, stronger temporal prediction, and domain adaptation.

\section*{Acknowledgements}
This work was supported by Institute of Information \& Communications Technology Planning \& Evaluation (IITP) grant funded by the Korea government (MSIT) (No.2021-0-01364, An intelligent system for 24/7 real-time traffic surveillance on edge devices)

%
%
\bibliographystyle{splncs04}
\bibliography{main}

\clearpage \appendix
\renewcommand{\thepage}{S\arabic{page}}
\setcounter{page}{1}
\title{Syn2RealTrack: Bridging the Gap Between Synthetic and Real-World Datasets for Online Multi-View Multi-Target Tracking}
\titlerunning{Syn2RealTrack: Online MTMC Tracking (Supplementary)}
\author{Supplementary Material}
\authorrunning{Supplementary Material}
\institute{ }
\maketitle

The supplementary material is organized as follows: \\
$\triangleright$~\cref{sec:2D_object_detection_details}: 2D Object Detection. \\
$\triangleright$~\cref{sec:re_id_pose_estimation_details}: Re-Identification and Pose Estimation. \\
$\triangleright$~\cref{sec:2D_tracking_details}: 2D Single-View Tracking. \\
$\triangleright$~\cref{sec:point_cloud_construction_details}: Point Cloud Construction.

\section{2D Object Detection}
\label{sec:2D_object_detection_details}

\subsection{Dataset Preprocessing}

The detection ground truth provides both 3D bounding box annotations and their corresponding 2D bounding box annotations. Since the number of 2D bounding boxes is relatively large, directly training on all available annotations may require a longer convergence time, and using the entire training and validation sets without scenario-specific selection may reduce the model's generalization to the test scenarios. The raw annotations also include redundant, small, or invalid 2D bounding boxes, which may increase false-positive detections. To address these issues, we apply three preprocessing steps before training: lookup-table mapping that summarizes per-object-type shape statistics for fallback object typing, scenario-specific selection that matches the training data to each test scenario, and filtering of redundant, small, or invalid 2D bounding boxes. Each step is described as follows:

\noindent\textbf{Lookup Table Mapping}
From the provided ground-truth annotations, we derive per-object-type shape statistics by measuring the minimum, maximum, and mean size of every object category. On this basis, objects are grouped into three types:
\begin{itemize}
  \item \textbf{Static Objects}: objects that stay in place for the entire observation period.
  \item \textbf{Fixed-Shape Objects}: objects whose shape stays constant whether they are moving or at rest.
  \item \textbf{Dynamic-Shape Objects}: objects whose shape changes as they move.
\end{itemize}
These statistics allow the framework to seed each object with a default shape, so that the evaluation still yields meaningful results even when later matching steps fail.

\noindent\textbf{Scenario-Specific Training-Set Selection}:
The framework aims to reduce the number of training bounding boxes while preserving detection accuracy as much as possible. Accordingly, we select the training data according to the specific conditions of each test scenario, which shortens training without sacrificing reliability. Specific scenarios are also selected to include object categories that are absent from other scenes, such as `AgilityDigit' and `FourierGR1T2'. For real-world scenes, beyond the AI City training dataset~\cite{tang_9th_2025}, the MTMMC dataset~\cite{woo_mtmmc_2024} is used to increase the amount and diversity of training data.

\begin{figure}[t]
	\centering
	\resizebox{\linewidth}{!}{
		\begin{tabular}{cc}
			\includegraphics[width=0.5\linewidth]{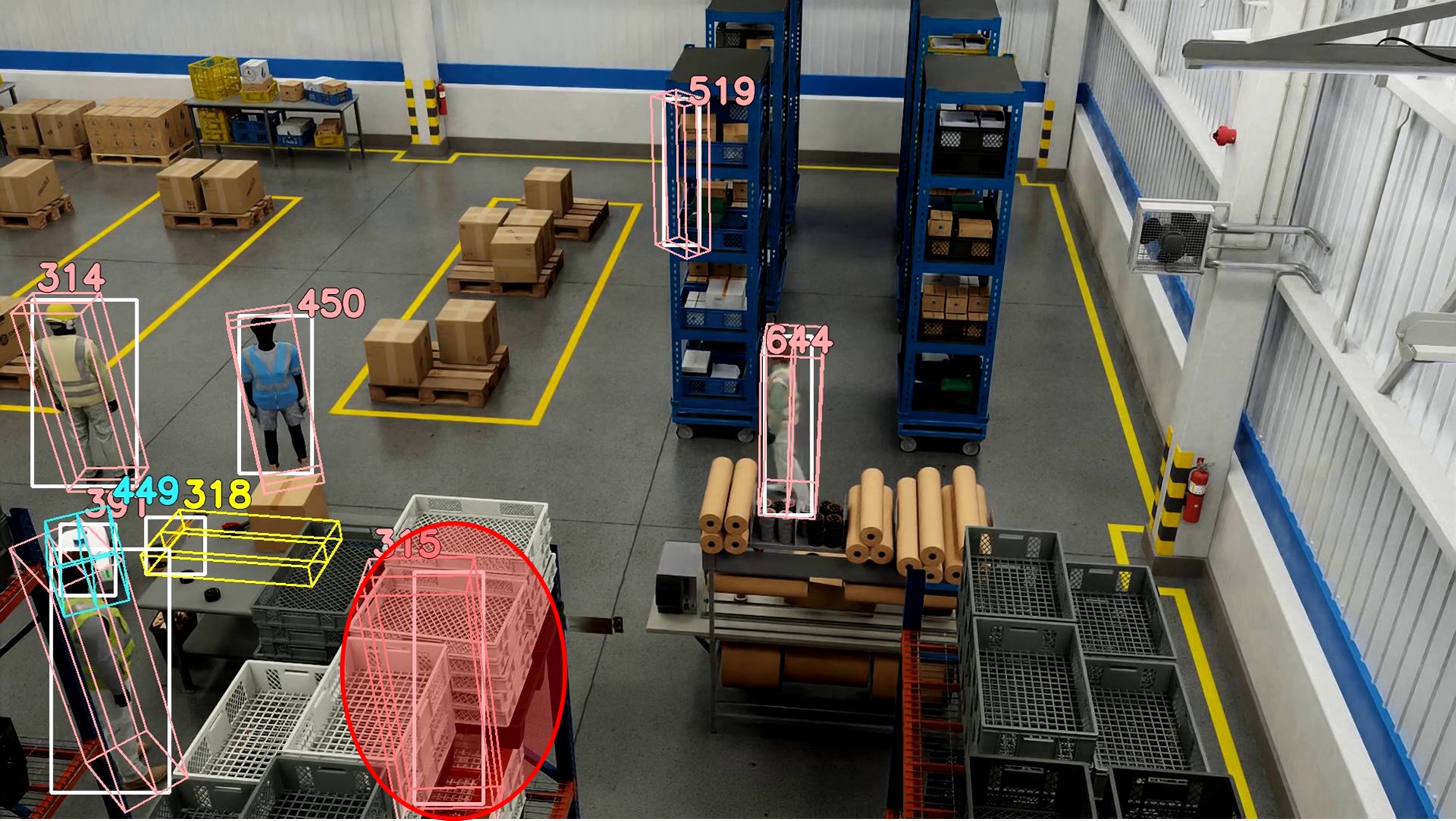} & \includegraphics[width=0.5\linewidth]{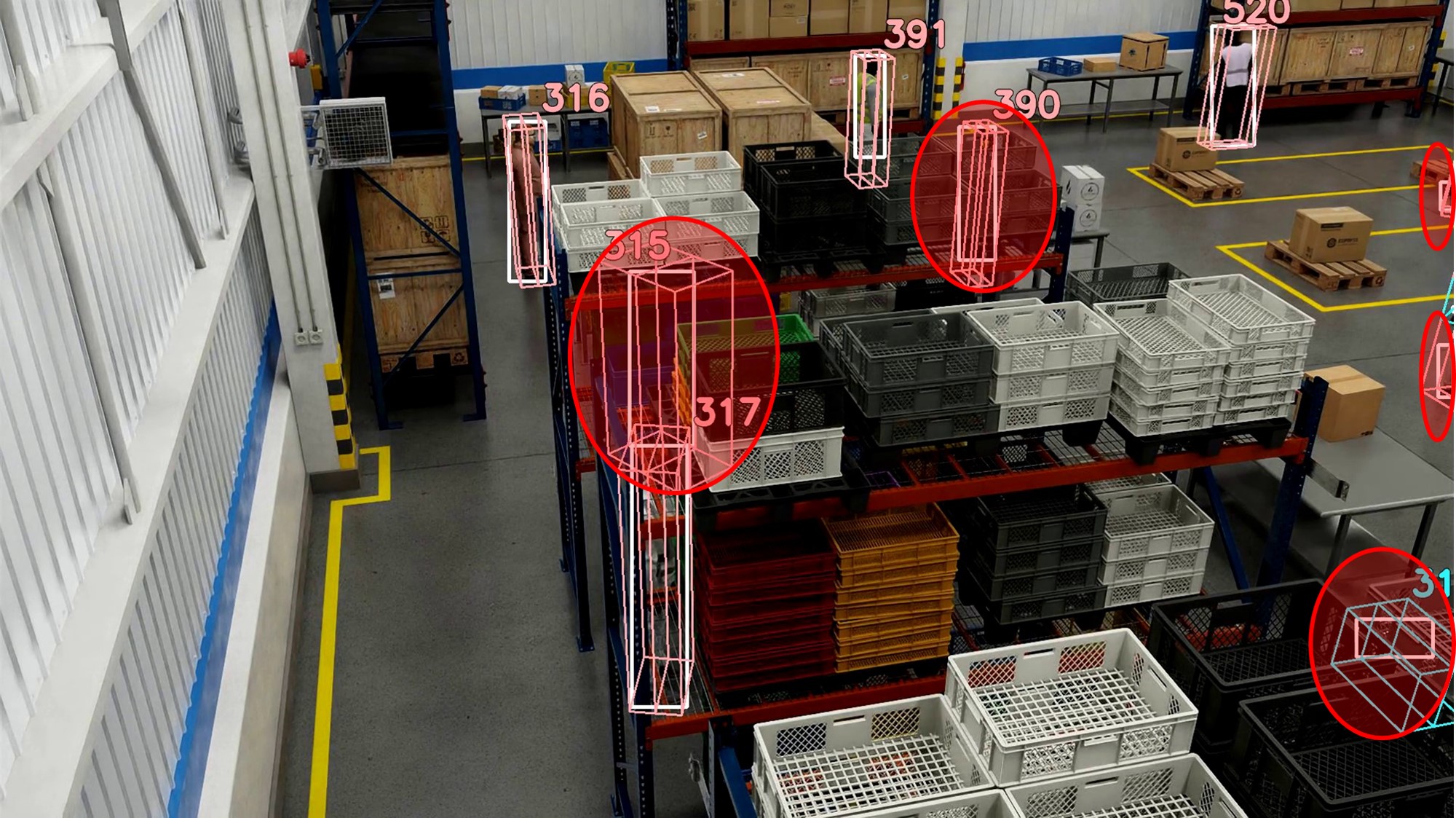}   \\
			\multicolumn{2}{c}{Scene without filtering} \\
			\includegraphics[width=0.5\linewidth]{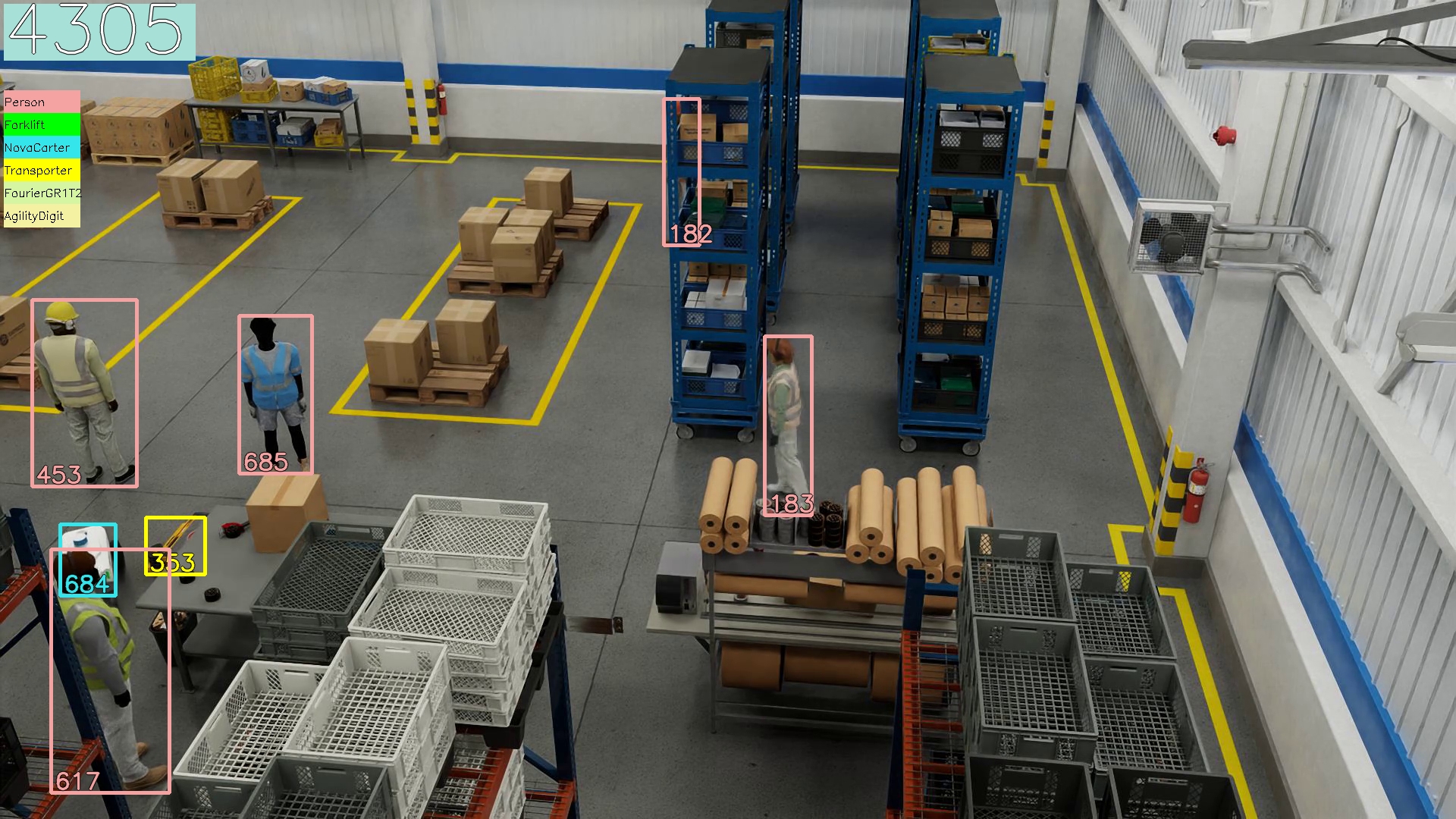} & \includegraphics[width=0.5\linewidth]{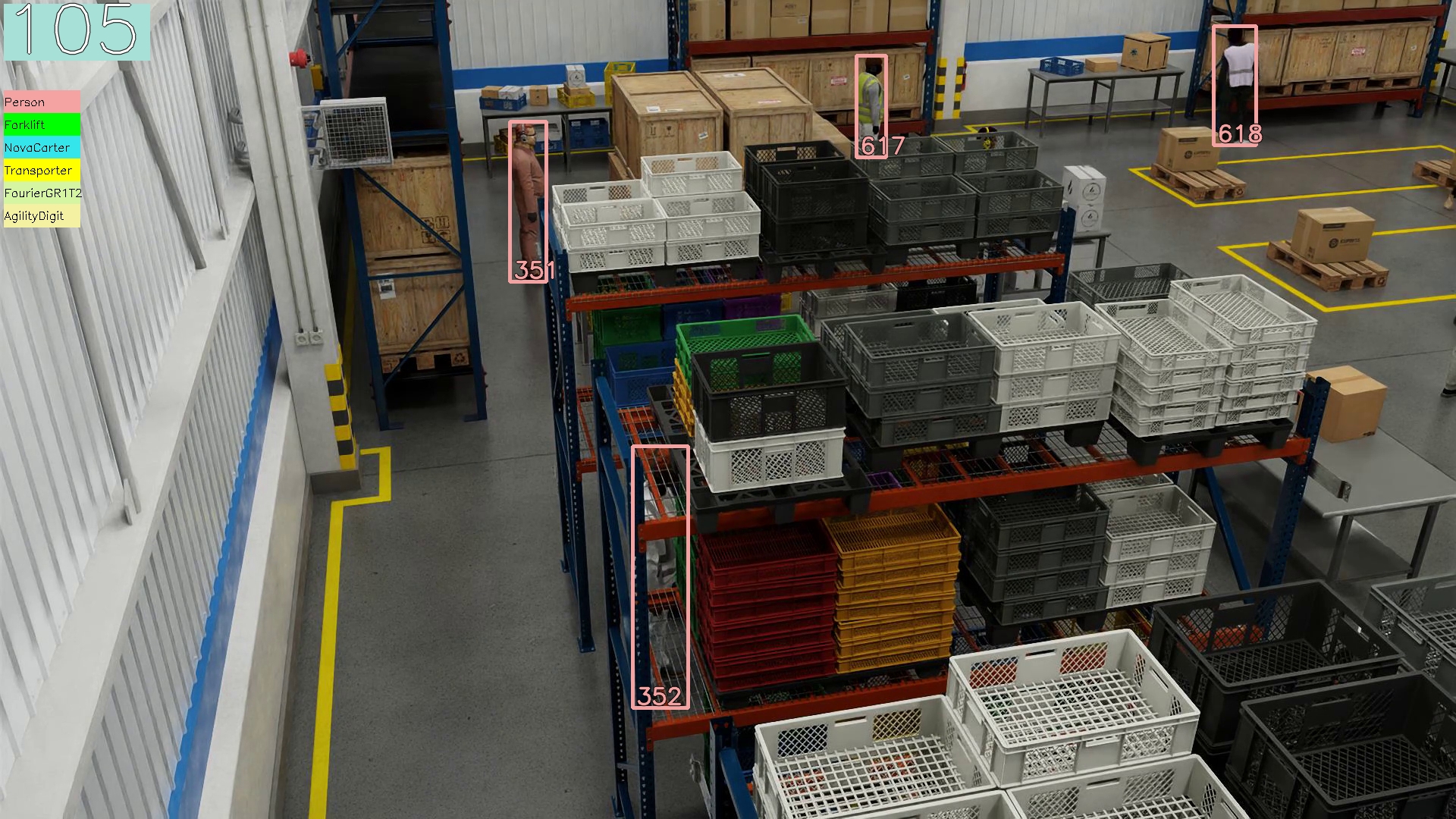}  \\
			\multicolumn{2}{c}{Scene with filtering}
		\end{tabular}
	}
	\caption{Visualization of the 2D bounding box filtering process. The bounding boxes removed from the training annotations are indicated by red ovals.}
	\label{fig:visualization_filtering_bbox}
\end{figure}

\noindent\textbf{Filtering Small or Invalid 2D Bounding Boxes}:
Redundant bounding boxes, as illustrated in~\cref{fig:visualization_filtering_bbox}, may increase the likelihood of false-positive detections and consequently degrade the reliability of the detection results. To mitigate this issue, we filter out irrelevant annotations: bounding boxes with excessively small spatial extent, as well as those covering unseen or invalid objects, are removed from the training annotations.

\noindent\textbf{2D Object Detection Training Step}:
We adopt RF-DETR~\cite{robinson2026rfdetr}, a real-time detection transformer, as the object detector. Specifically, we train the 2x-large variant for 200 epochs on input images with a resolution of $1920$~pixels and a batch size of 4; this setting balances memory usage and training stability.

\begin{figure}[t]
	\centering
	\resizebox{\linewidth}{!}{
		\begin{tabular}{cc}
			\includegraphics[width=0.5\linewidth]{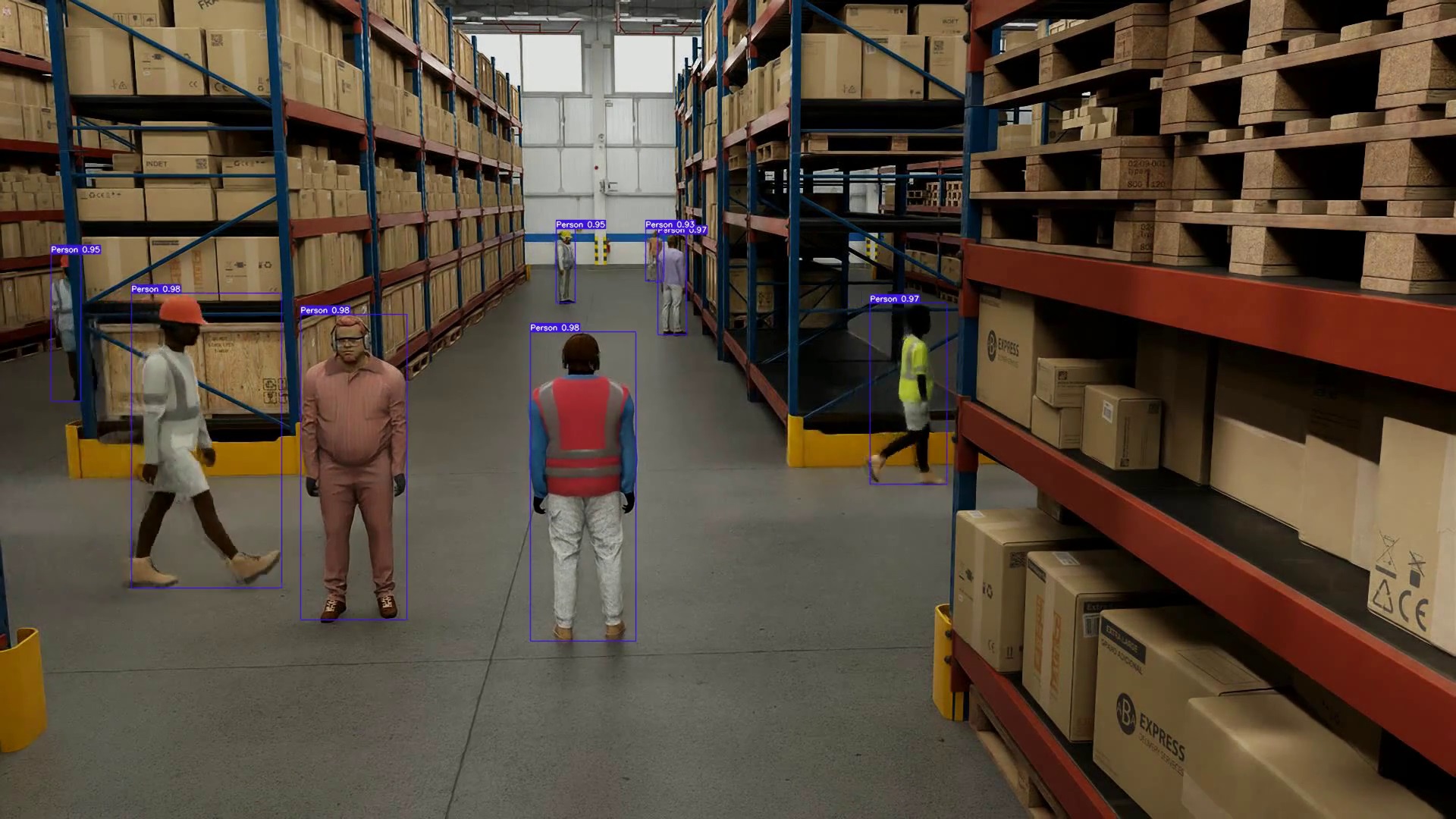} & \includegraphics[width=0.5\linewidth]{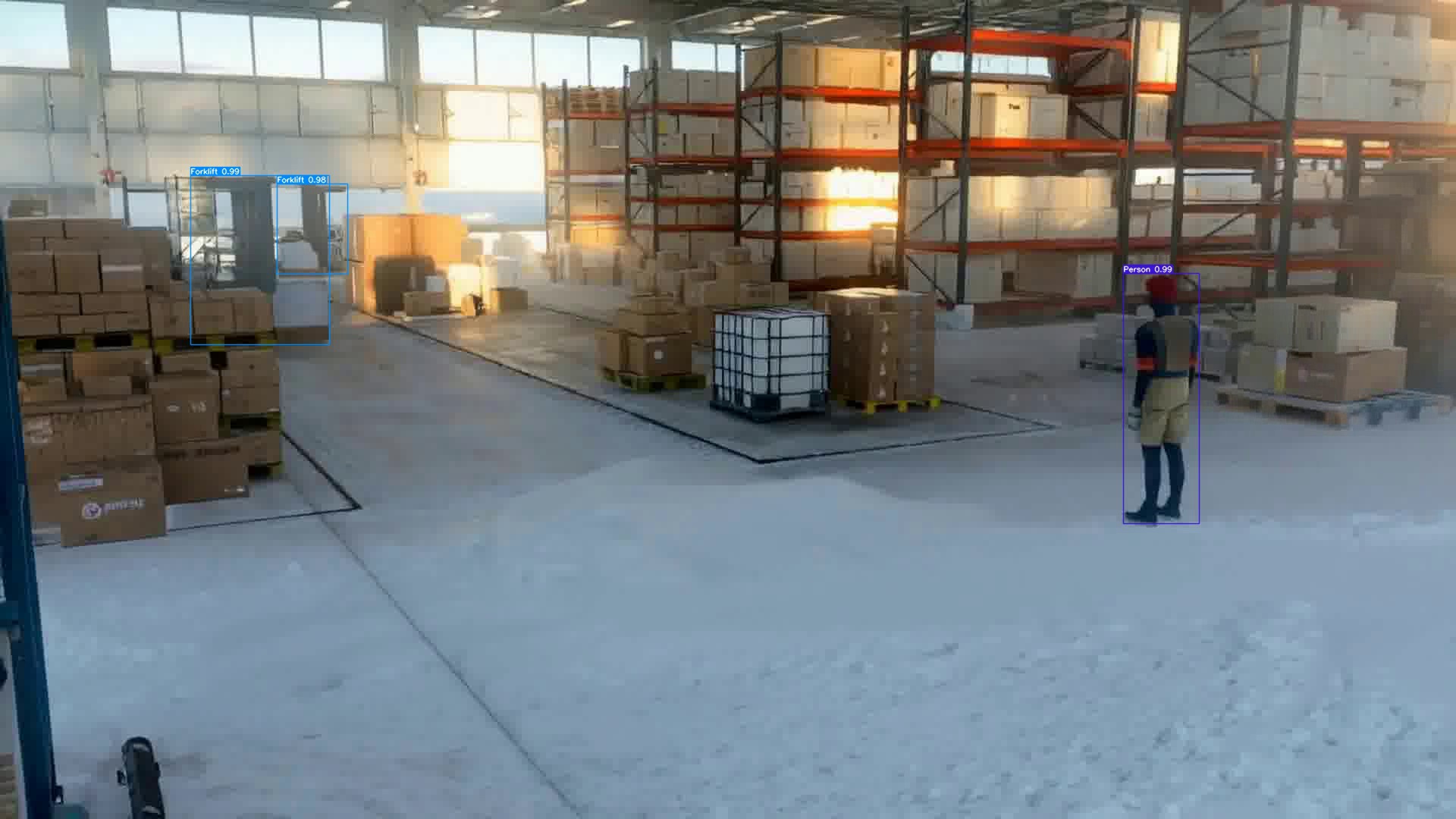}   \\
			\includegraphics[width=0.5\linewidth]{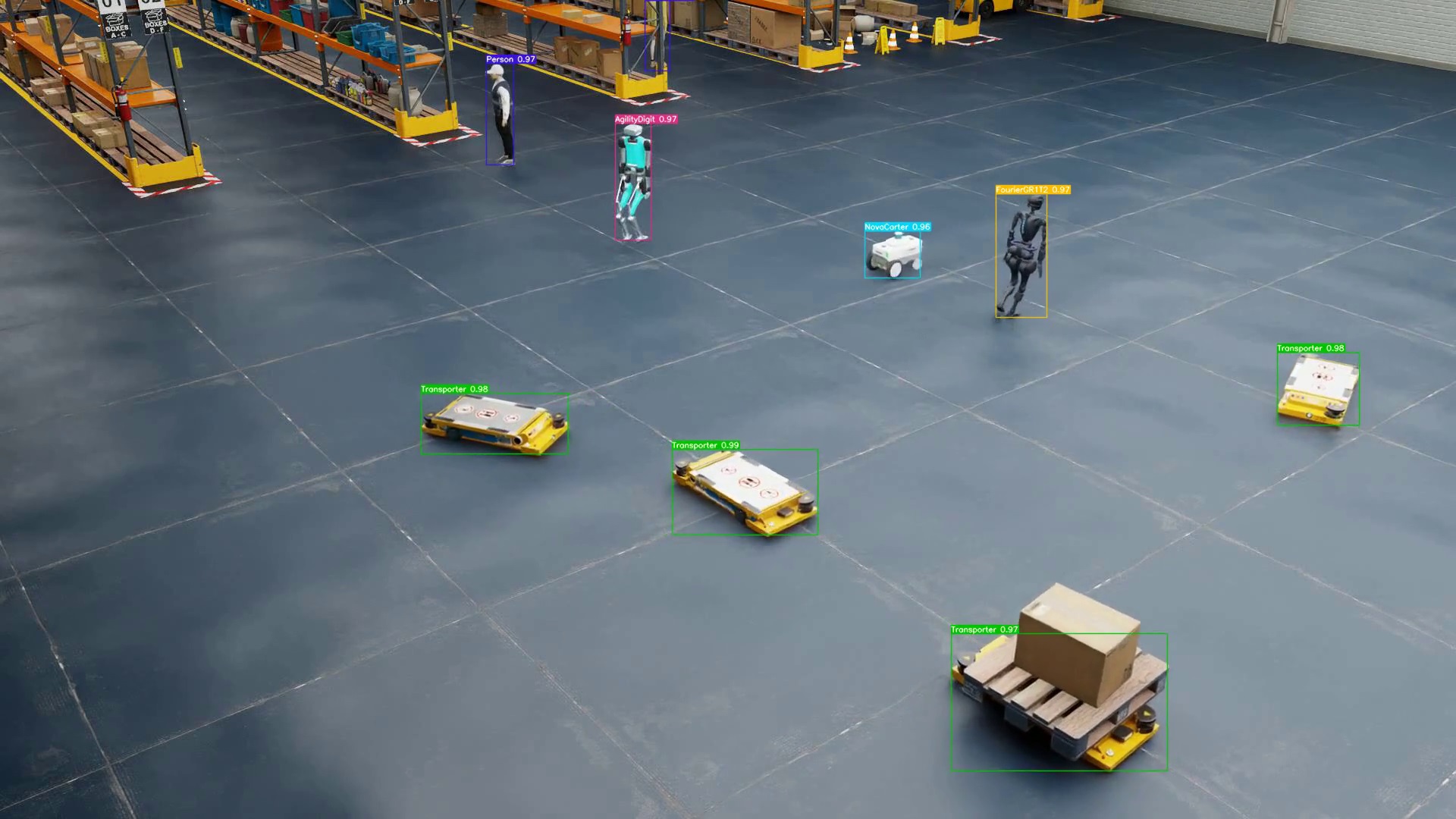} & \includegraphics[width=0.5\linewidth]{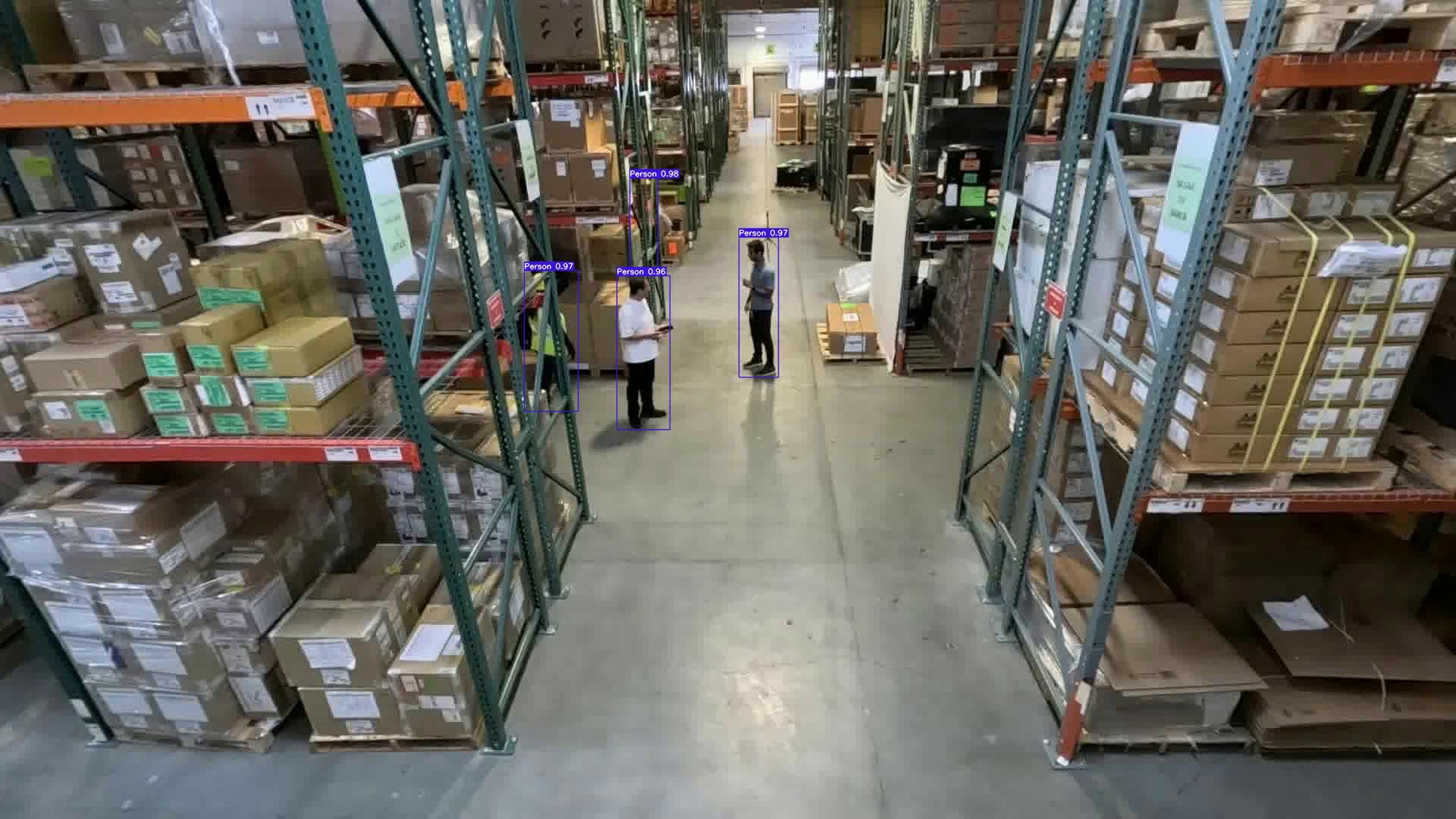} \\
			\multicolumn{2}{c}{\includegraphics[width=0.5\linewidth]{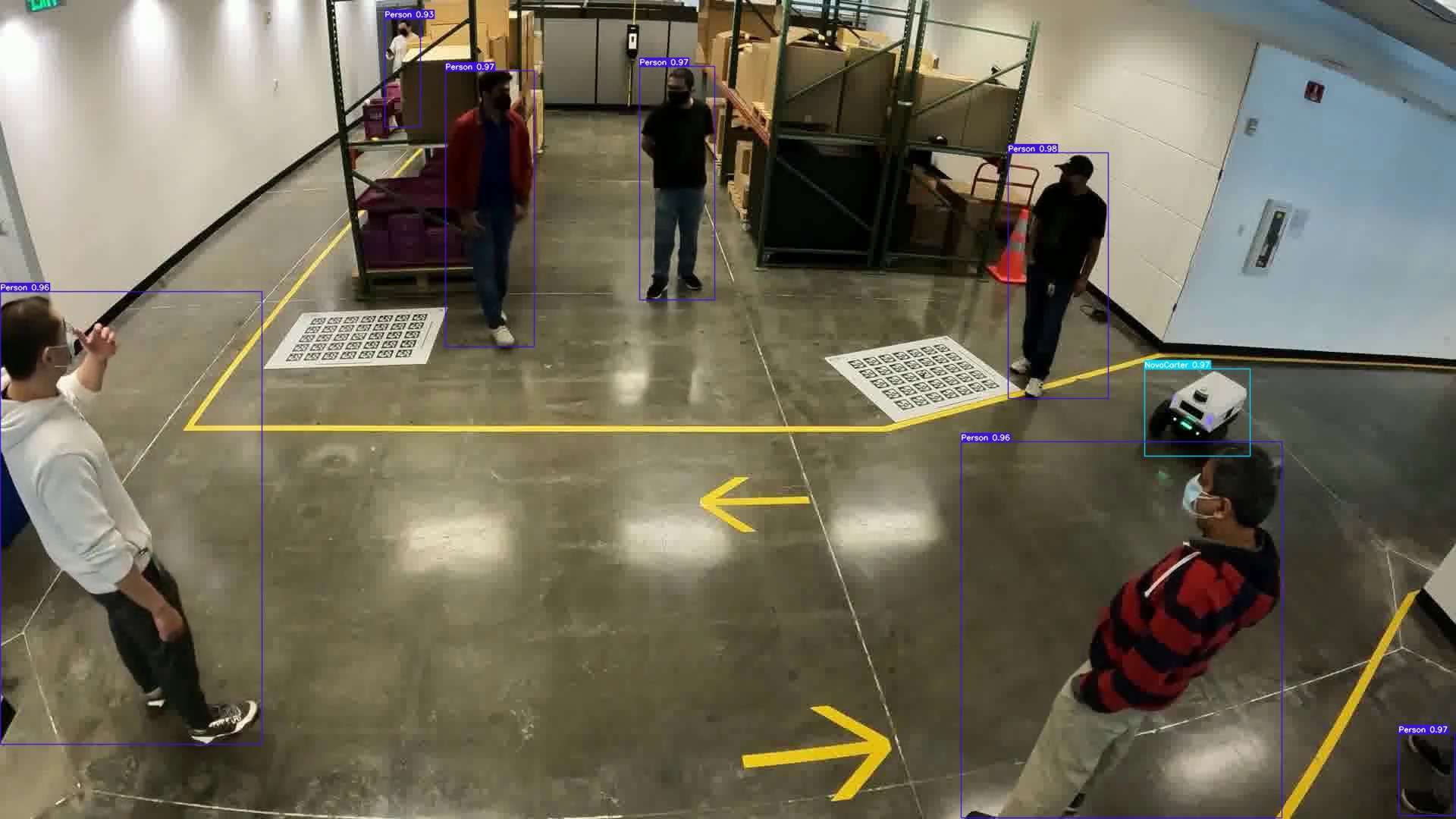}}
		\end{tabular}
	}
	\caption{Qualitative 2D detection results on scenes Warehouse~023--027 of the AI City Challenge 2026 Track~1 test set.}
	\label{fig:2d_detection_visualization}
\end{figure}

\noindent\textbf{2D Object Detection Inference Step}:
During inference, the trained RF-DETR model processes images at the same $1920$-pixel resolution with a batch size of 16 to improve throughput. We perform non-maximum suppression at an Intersection over Union (IoU) threshold of 0.5 to reduce redundant overlapping bounding boxes and discard detections with confidence scores below 0.1. Qualitative detection results for the five test scenes are shown in~\cref{fig:2d_detection_visualization}.

\section{Re-Identification and Pose Estimation}
\label{sec:re_id_pose_estimation_details}

\begin{figure}[t]
	\centering
	\resizebox{\linewidth}{!}{
		\begin{tabular}{cc}
			\includegraphics[width=0.5\linewidth]{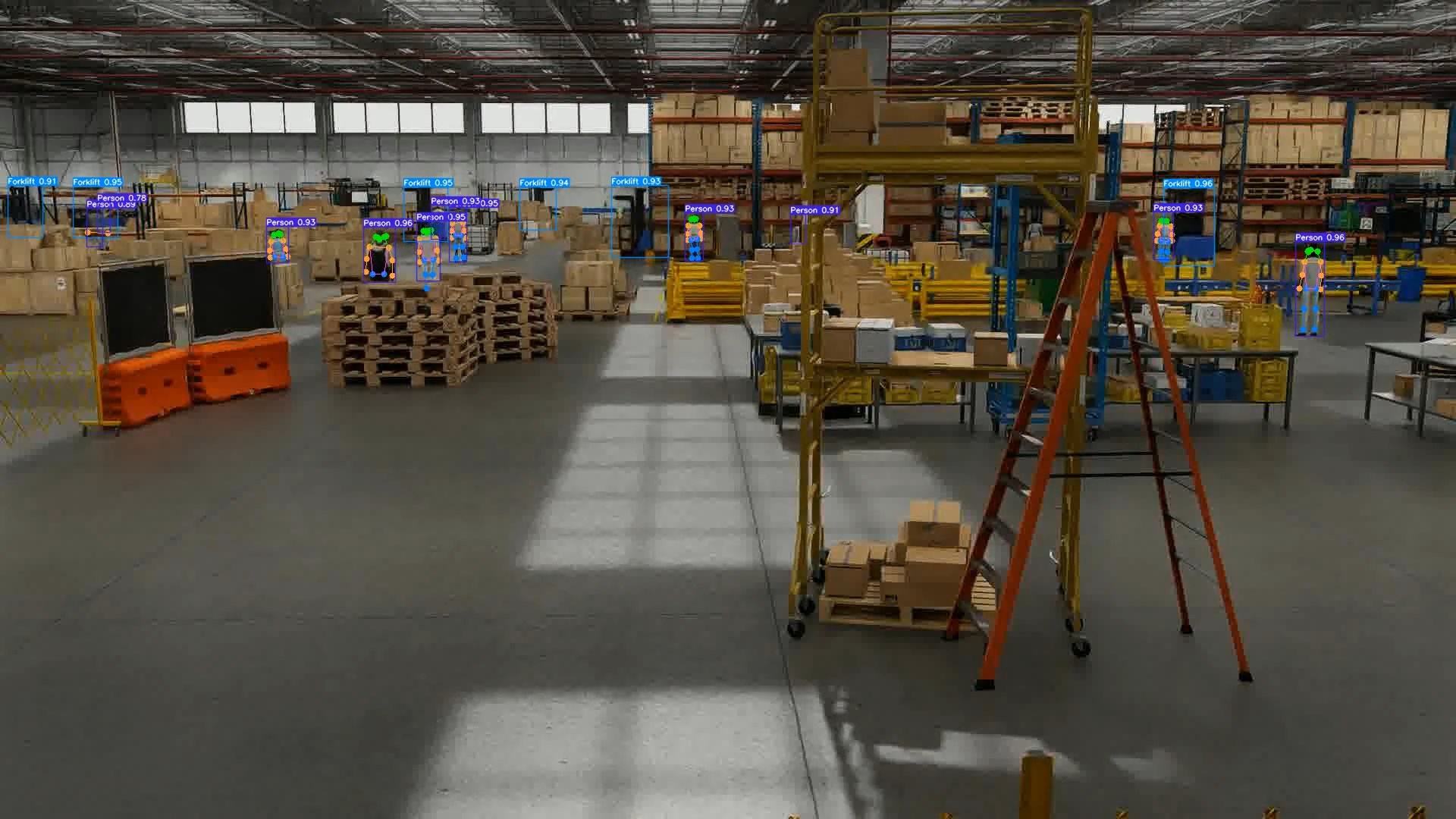} & \includegraphics[width=0.5\linewidth]{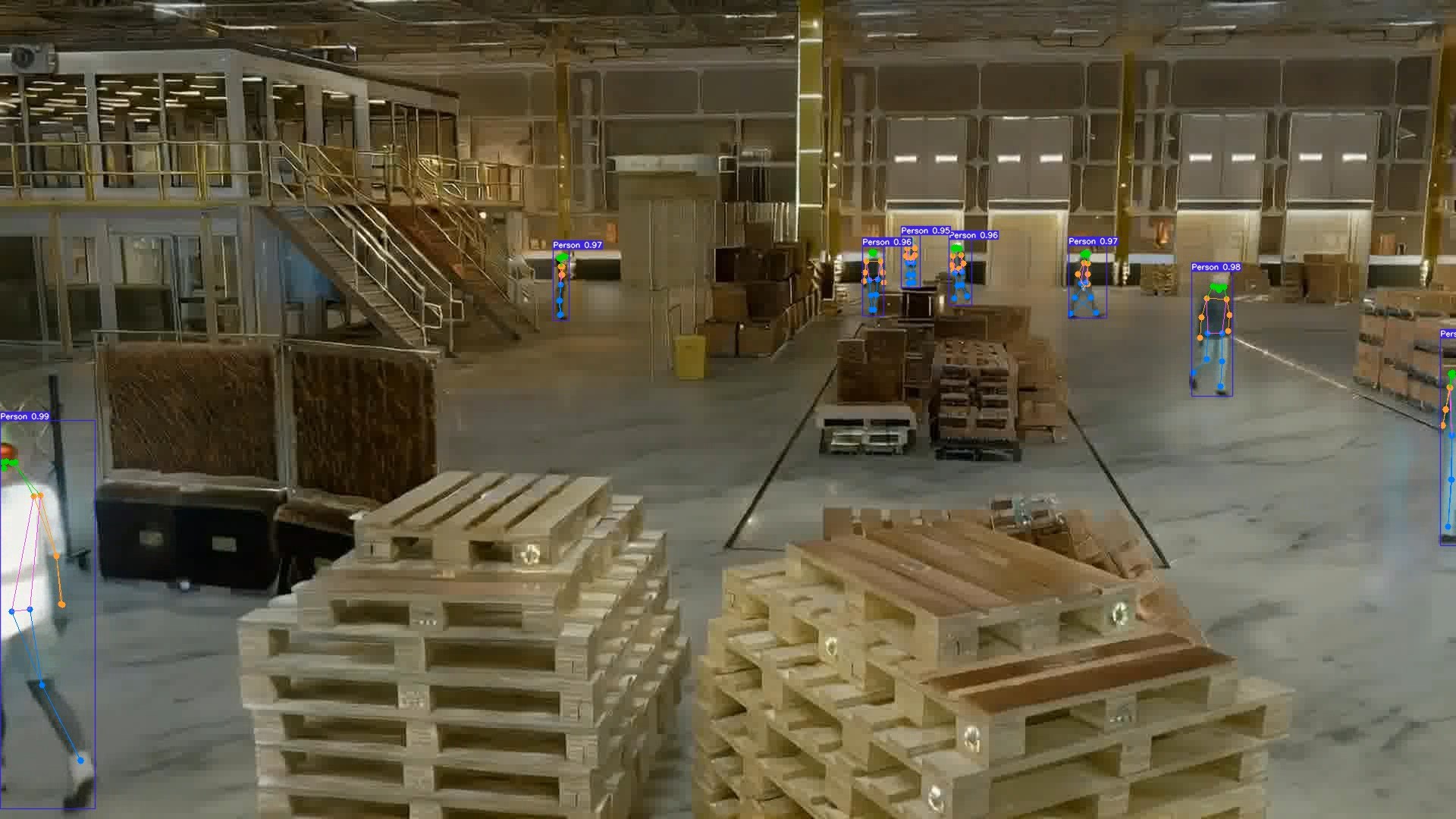}   \\
			\includegraphics[width=0.5\linewidth]{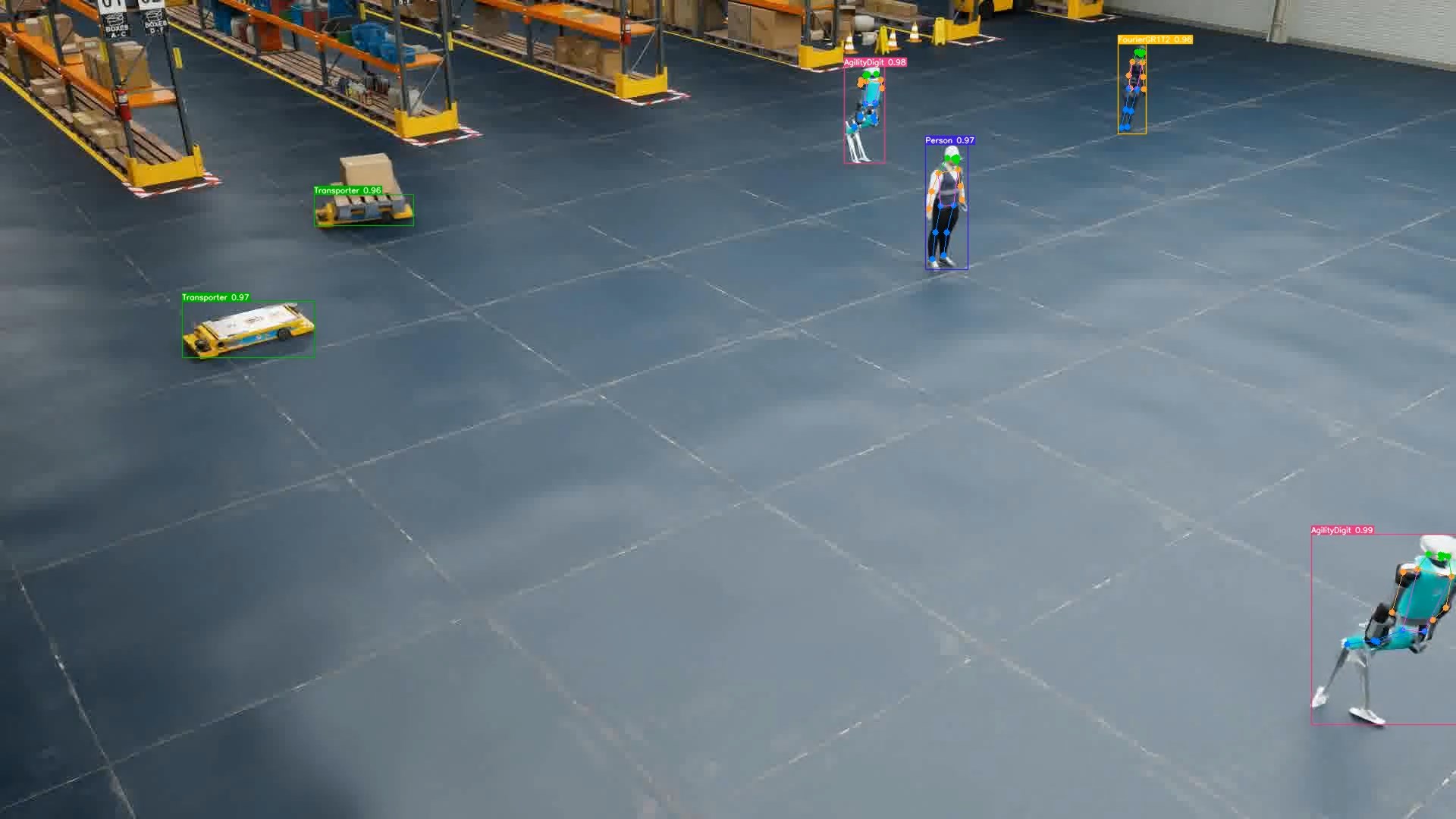} & \includegraphics[width=0.5\linewidth]{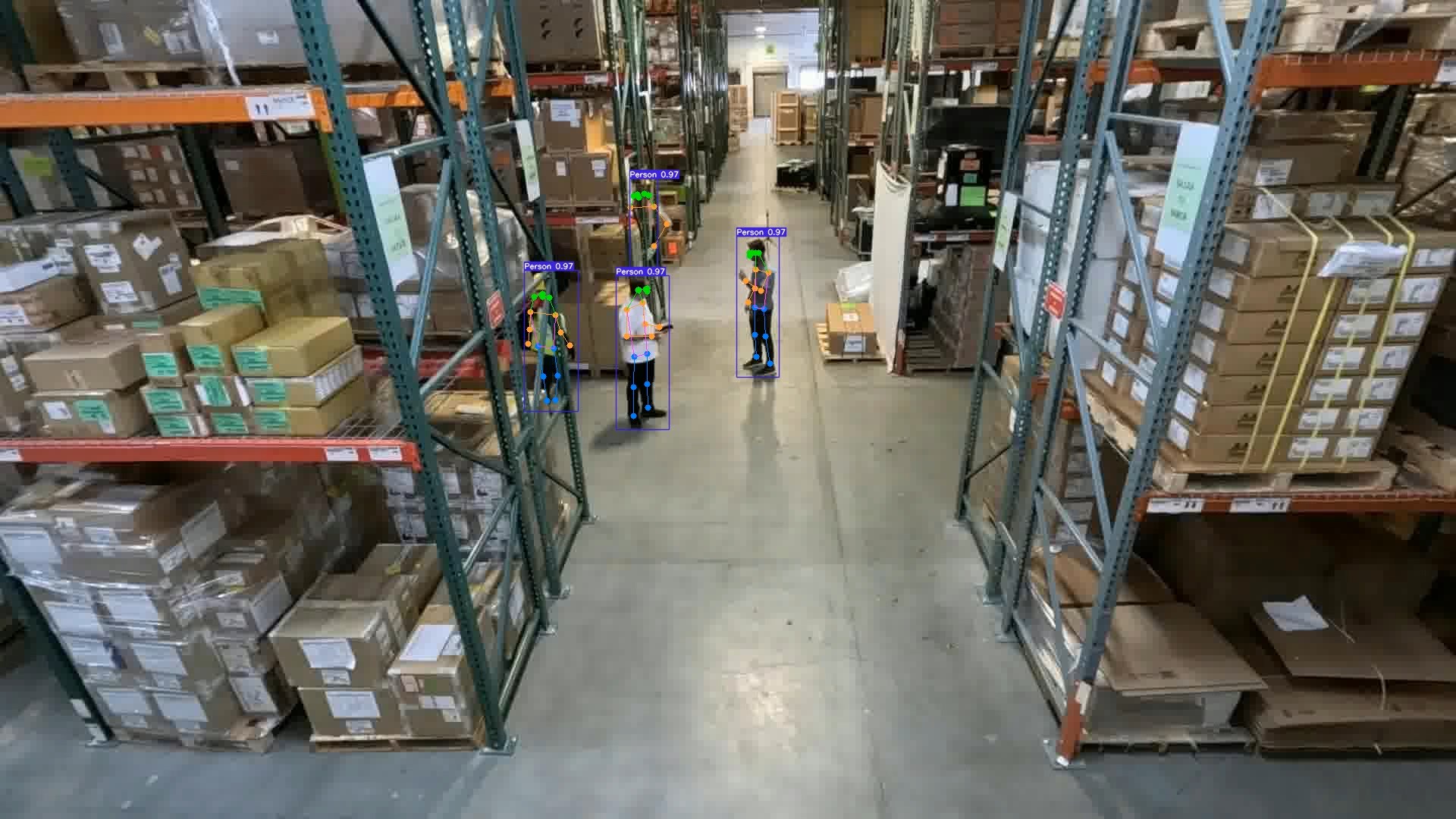} \\
			\multicolumn{2}{c}{\includegraphics[width=0.5\linewidth]{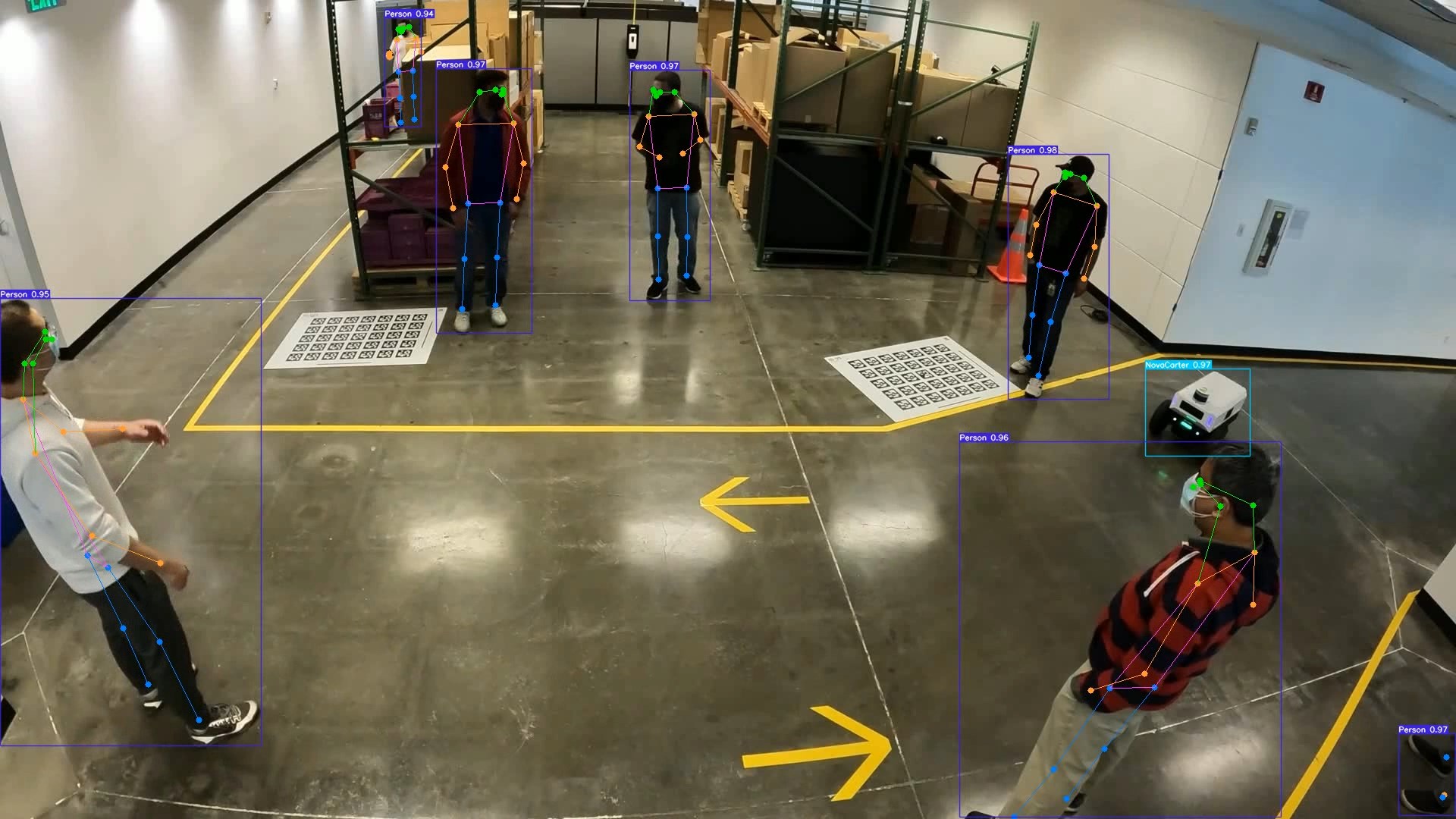}}
		\end{tabular}
	}
	\caption{Qualitative pose-estimation results (ViTPose++~\cite{xu_vitpose_2024} keypoint overlays) on scenes Warehouse~023--027.}
	\label{fig:2d_pose_estimation_visualization}
\end{figure}

\noindent\textbf{Pose Estimation} We use the pretrained ViTPose++ model~\cite{xu_vitpose_2024} for pose estimation to extract keypoints from detected person regions (as shown in~\cref{fig:2d_pose_estimation_visualization}). ViTPose++ adopts a plain, non-hierarchical vision transformer encoder with a lightweight keypoint decoder, which localizes body keypoints effectively with a simple, scalable architecture. In addition, the model introduces knowledge factorization through task-agnostic and task-specific feed-forward networks, which allows it to handle heterogeneous body keypoint categories across different pose estimation tasks. The pretrained model supplies the pose information used by the subsequent feature extraction, association, and refinement steps.

\noindent\textbf{Re-Identification} For person re-identification (Re-ID), which relies on deep visual features, we adopt Keypoint Promptable Re-Identification (KPR)~\cite{DBLP:conf/eccv/SomersAV24} for feature extraction. KPR uses pose keypoints to divide the human body into six regions and extracts region-level features from each part. This design emphasizes human-centric cues, reduces background interference, and improves robustness under occlusion. Although KPR supports negative keypoints for modeling occluded or unreliable body regions, only positive keypoints are used during training and inference in our pipeline. Since identity labels differ across scenes in the AI City 2026 Track 1 dataset, the training and validation sets are merged and relabeled into a unified dataset containing 155 identities. The model is trained for 110 epochs, with a total training time of approximately 3 hours and 7 minutes.

\section{2D Single-View Tracking}
\label{sec:2D_tracking_details}

\begin{figure}[t]
	\centering
	\resizebox{\linewidth}{!}{
		\begin{tabular}{cc}
			\includegraphics[width=0.5\linewidth]{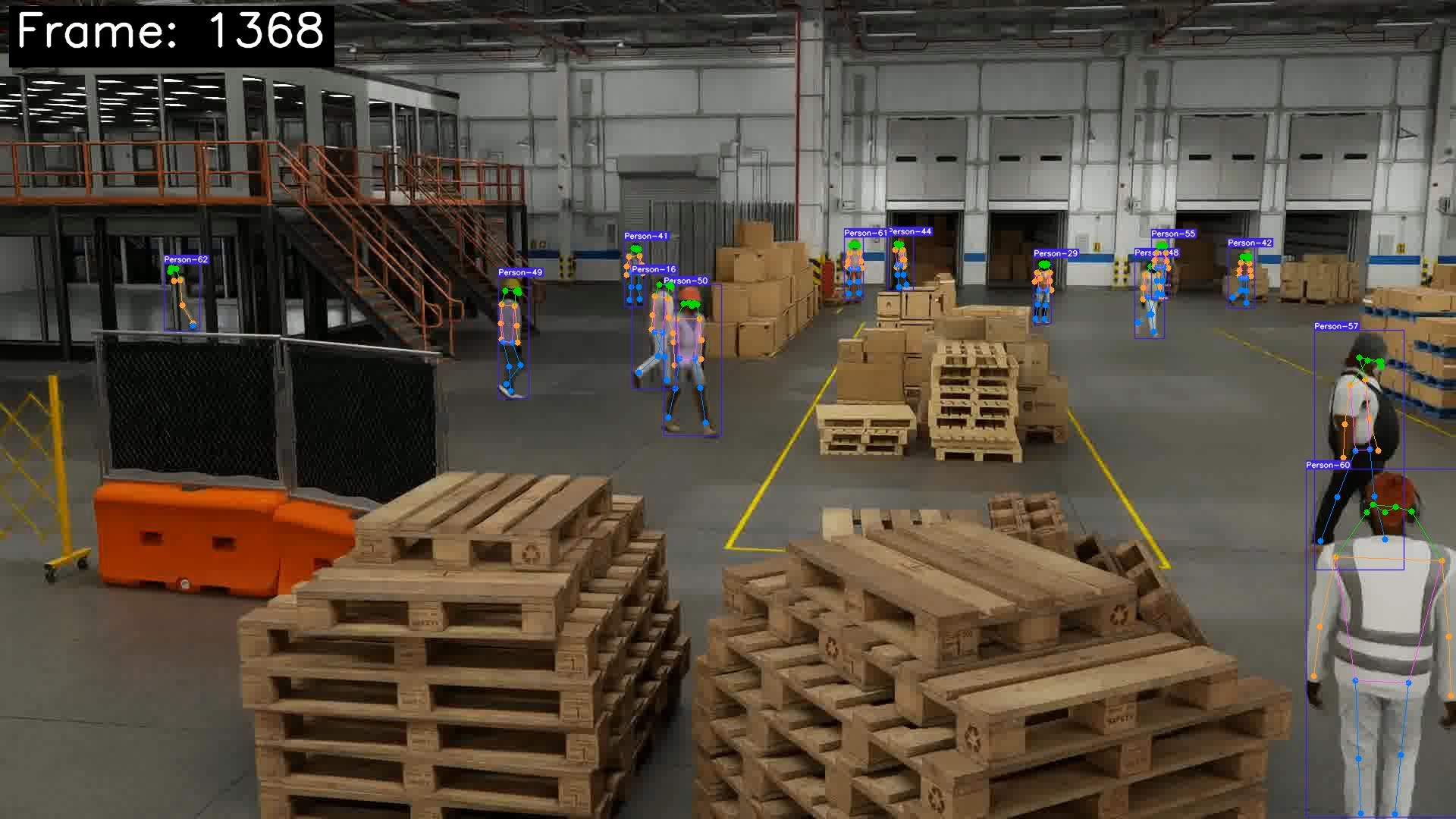} & \includegraphics[width=0.5\linewidth]{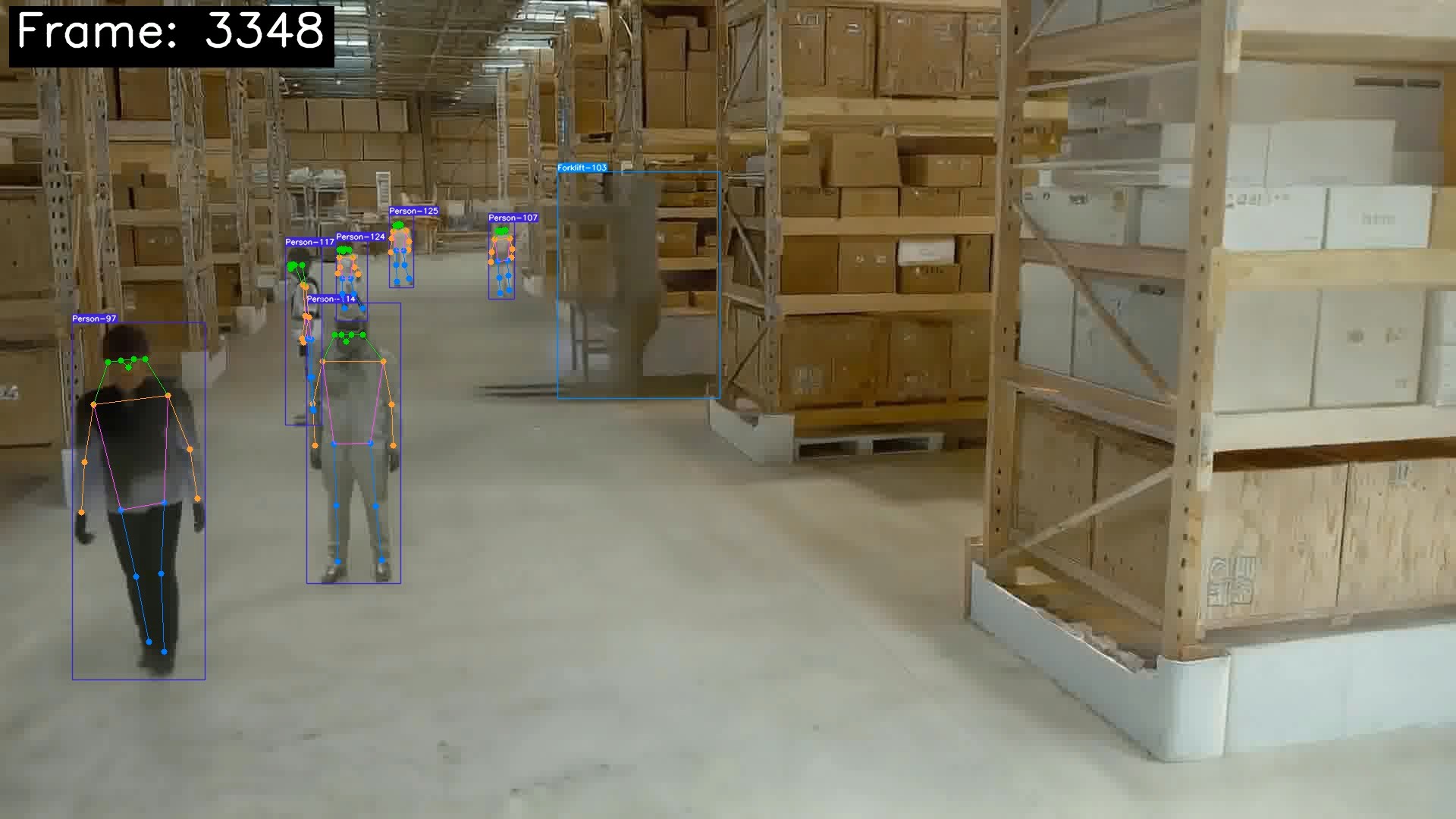}   \\
			\includegraphics[width=0.5\linewidth]{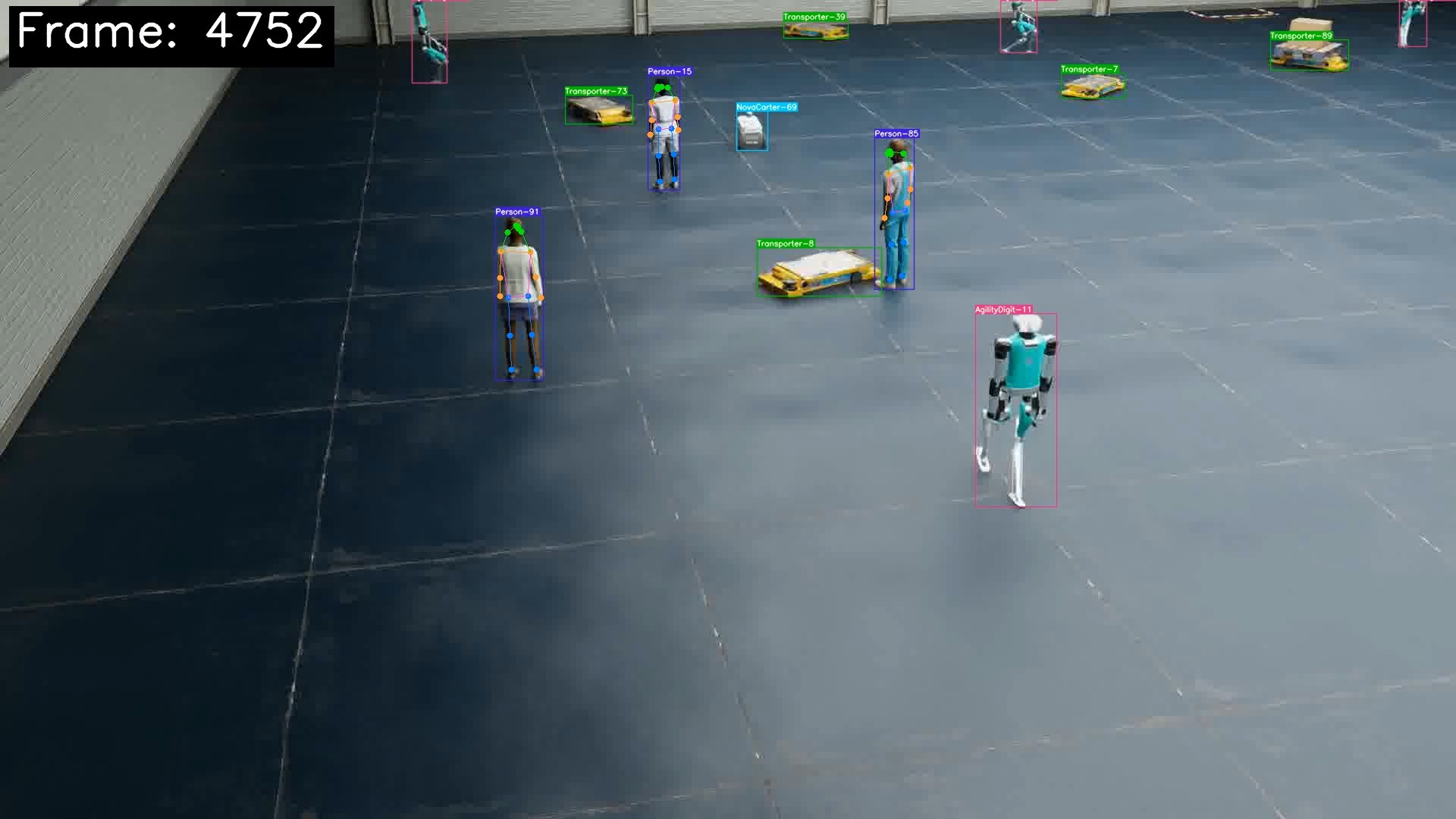} & \includegraphics[width=0.5\linewidth]{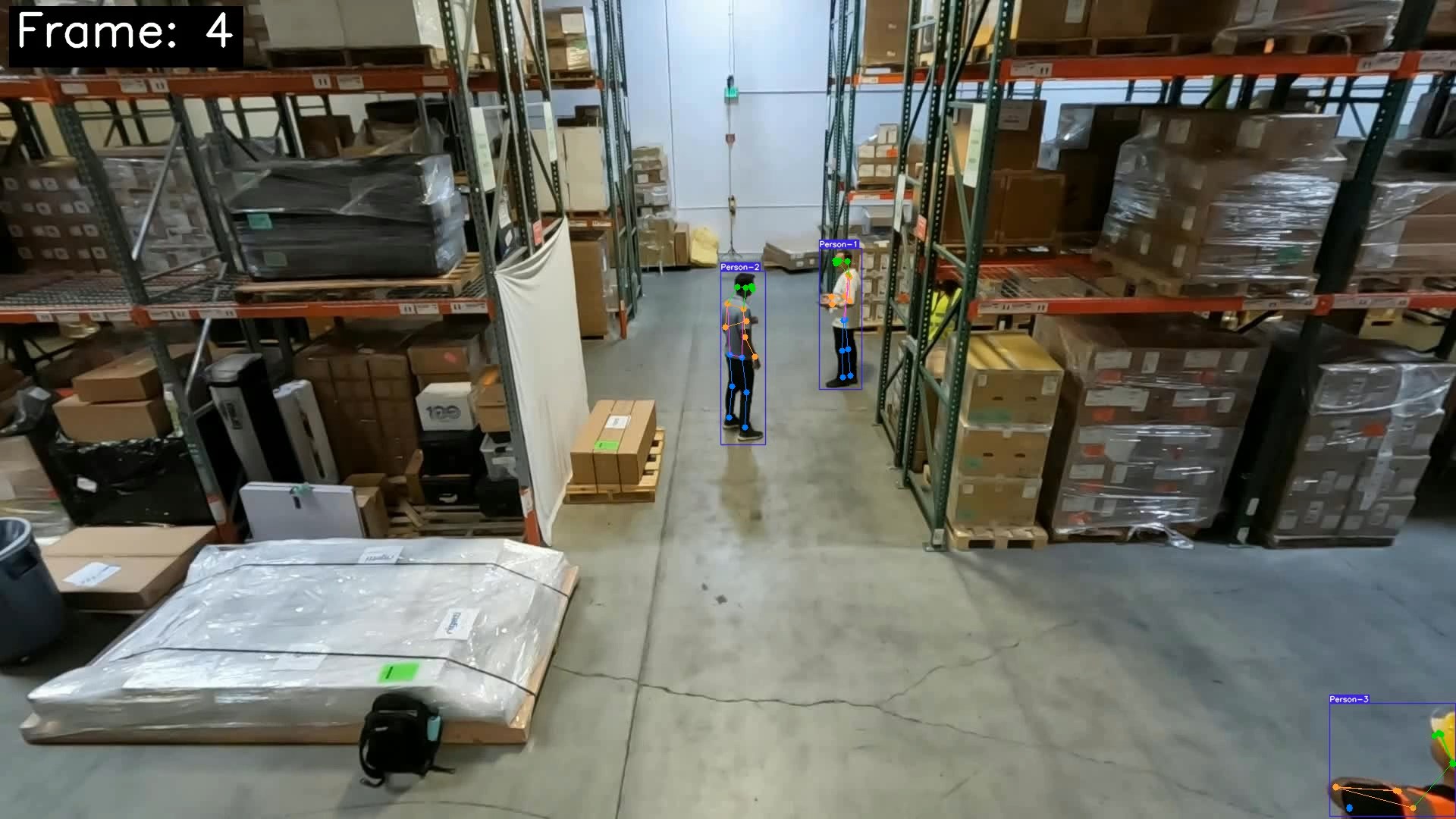} \\
			\multicolumn{2}{c}{\includegraphics[width=0.5\linewidth]{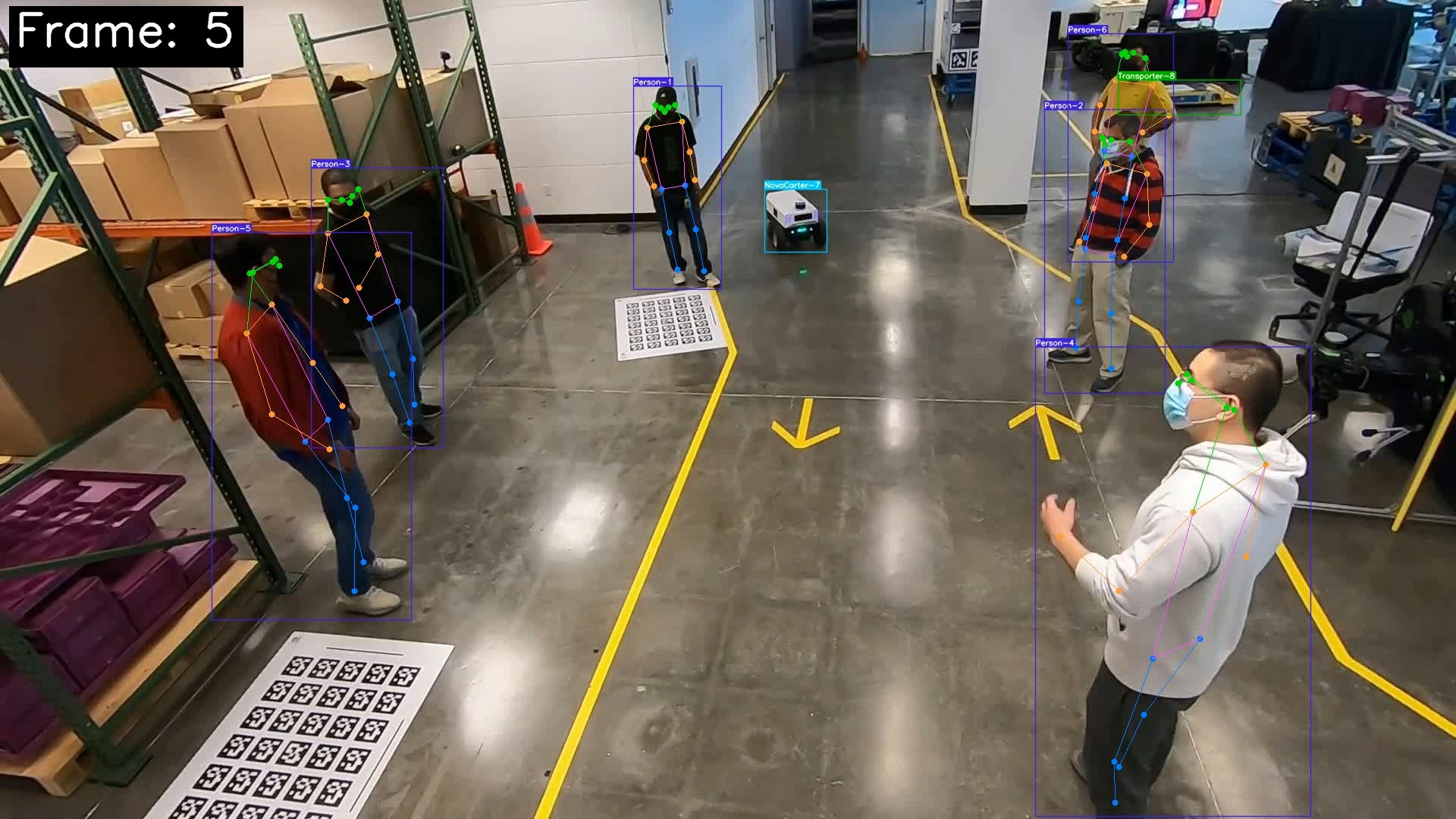}}
		\end{tabular}
	}
	\caption{Qualitative 2D single-view tracking results on scenes Warehouse~023--027; each tracked object carries a persistent identity.}
	\label{fig:2d_tracking_visualization}
\end{figure}

The result of 2D single-view tracking is visualized in~\cref{fig:2d_tracking_visualization}. In the shown frames, the method tracks multiple objects in a single camera view through occlusions and close object interactions: each object keeps its identity across frames, and the bounding boxes follow the objects throughout the sequence.

\section{Point Cloud Construction}
\label{sec:point_cloud_construction_details}

\begin{figure}[t]
    \centering
    \begin{minipage}{0.48\linewidth}
        \centering
        \includegraphics[width=\linewidth]{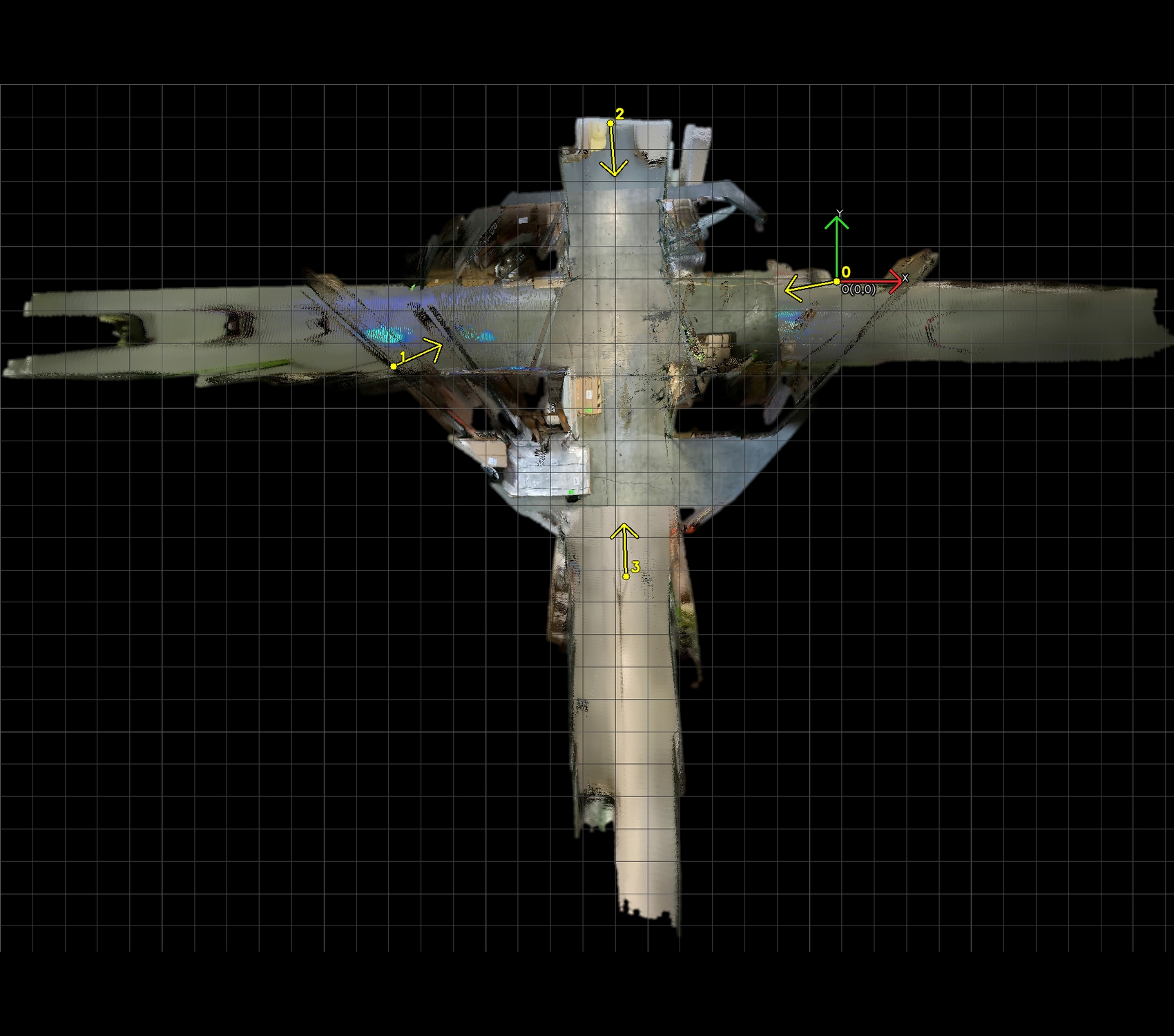}\\[-1mm]
        \small (a) Warehouse~026
    \end{minipage}
    \hfill
    \begin{minipage}{0.48\linewidth}
        \centering
        \includegraphics[width=\linewidth]{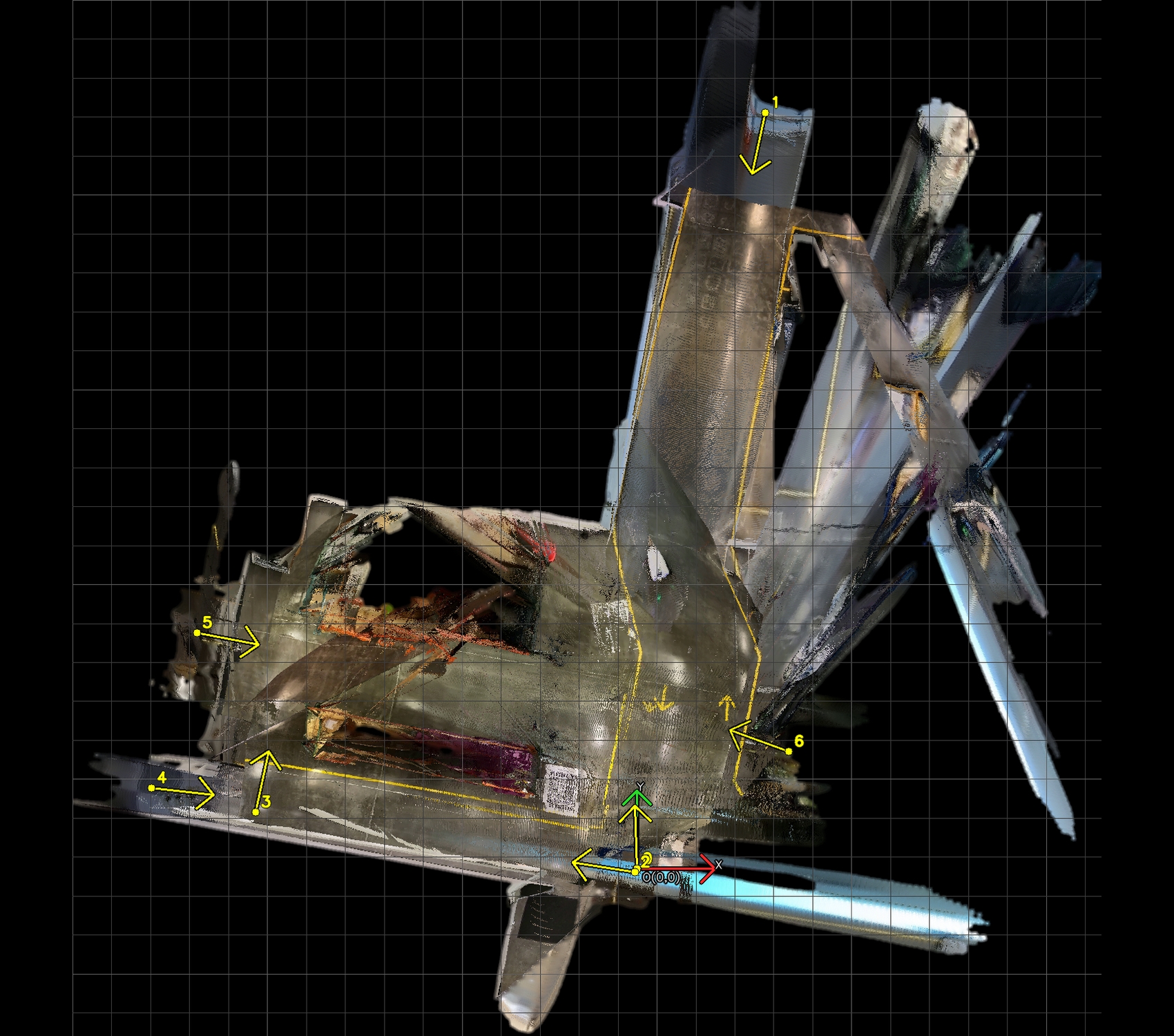}\\[-1mm]
        \small (b) Warehouse~027
    \end{minipage}
    \caption{Scene-level BEV visualizations constructed from DA3-estimated point clouds for (a) Warehouse~026 and (b) Warehouse~027.}
    \label{fig:bev_realworld_pointclouds}
	\vspace{-0.3cm}
\end{figure}

Since depth maps are unavailable at test time, we generate point-cloud guidance using Depth Anything~3 (DA3)~\cite{lin2026depth} in its pose-conditioned multi-view setting. Specifically, we adopt the updated DA3 Nested Giant-Large 1.1 checkpoint, which integrates any-view geometry with metric-depth scaling. At each synchronized time step, images from all cameras are processed together with the provided intrinsic and extrinsic parameters, $K$ and $E$. The resulting depth estimates are back-projected into a shared world coordinate system and merged to form a colored point cloud. \cref{fig:da3_multiview_pointcloud} presents the multi-view RGB inputs alongside their DA3-estimated point clouds, while~\cref{fig:bev_realworld_pointclouds} illustrates the corresponding BEV projections for two real-world scenes.

\end{document}